\documentclass[a4paper,fleqn]{cas-sc}

\usepackage[authoryear,longnamesfirst]{natbib}

\usepackage{subcaption}
\usepackage{tabularx}

\usepackage{amsmath} 
\usepackage{fontspec}
\usepackage{tipa}
\newfontfamily\assamese[Script=Bengali,Language=Assamese,Path=./]{Lohit-Assamese.ttf}

\newfontfamily\hindi[Script=Devanagari,Language=Hindi,Path=./]{NotoSansDevanagari-Regular.ttf}
\def\tsc#1{\csdef{#1}{\textsc{\lowercase{#1}}\xspace}}
\tsc{WGM}
\tsc{QE}

\begin{document}
\let\WriteBookmarks\relax
\def\floatpagepagefraction{1}
\def\textpagefraction{.001}

\shorttitle{}    

\shortauthors{}  



\title [mode = title]{Robust Assamese Speech Recognition through Controlled Fine-Tuning of Whisper Models}

\tnotetext[1]{} 

\author[1]{Ganapati Das}[orcid=0009-0008-2378-027X]
\fnmark[1]
\ead{ganapatidas@gauhati.ac.in}
\credit{Conceptualization, Methodology, Writing - original draft}

\author[1]{Dwipen Laskar} [orcid=0000-0001-6081-4338]
\cormark[1]
\fnmark[1]
\ead{laskardwipen@gauhati.ac.in}
\credit{Validation, Writing - review and editing}

\author[1]{Hasin Afzal Ahmed} [orcid=0009-0004-9070-3232]
\fnmark[1]
\ead{hasin@gauhati.ac.in}
\credit{Data curation, Software}

\author[1]{Sanjib Kr Kalita} [orcid=0000-0002-0576-3010]
\fnmark[1]
\ead{sanjib959@gauhati.ac.in}
\credit{Investigation, Visualization}

\author[2]{Kshirod Sarmah} [orcid=0009-0006-8780-5689]
\fnmark[1]
\ead{kshirodsarmah@gmail.com}
\credit{Resources, Validation}

\author[3]{Hem Chandra Das} [orcid=0000-0002-1142-4605]
\fnmark[1]
\ead{hemchandradas78@gmail.com}
\credit{Formal analysis, Writing - review and editing}

\author[4]{Manjula Kalita} [orcid=0009-0003-6170-9160]
\fnmark[1]
\ead{manjula_cse@gcuniversity.ac.in}
\credit{Writing - review and editing}


\affiliation[1]{organization={Department of Computer Science, Gauhati University},
                city={Guwahati},
                postcode={781014},
                state={Assam},
                country={India}}

\affiliation[2]{organization={Department of Computer Science, Pandit Deendayal Upadhyaya Adarsha Mahavidyalaya},
                city={Goalpara},
                postcode={783101},
                state={Assam},
                country={India}}

\affiliation[3]{organization={Department of Computer Science and Technology, Bodoland University},
                city={Kokrajhar},
                postcode={783370},
                state={Assam},
                country={India}}

\affiliation[4]{organization={Department of Computer Science and Engineering, Girijananda Chowdhury University},
                city={Guwahati},
                postcode={781017},
                state={Assam},
                country={India}}


\cortext[cor1]{Corresponding author}

\fntext[1]{These authors contributed equally to this work.}


















\begin{abstract}
Developing Automatic Speech Recognition (ASR) for morphologically rich, low-resource languages such as Assamese is challenging due to insufficient annotated speech data. The pretrained Whisper model performs poorly on Assamese speech recognition tasks. This paper presents a controlled, fine-tuned Whisper-based Assamese ASR system trained on the Mozilla Common Voice 24.0-Assamese corpus. A hardware-aware optimized training pipeline is implemented for resource-constrained environments, employing mixed-precision training and gradient accumulation on Tesla 4 Graphics Processing Units (T4 GPUs). The proposed fine-tuned model significantly outperformed the Zero-shot baseline, yielding Word Error Rate (WER), Character Error Rate (CER), Match Error Rate (MER), and Word Infomation Loss (WIL) of 43.17\%, 13.18\%, 43\%, and 64.81\%, respectively, achieving significant relative improvements of 78.26\%, 93.10\%, 57.0\%, and 35.19\% over the baseline. Semantic evaluation of the fine-tuned model also demonstrates notable improvement over a zero baseline, attaining Bilingual Evaluation Understudy (BLEU) and Metric for Evaluation of Translation with Explicit ORdering (METEOR) scores of 30.81 and 0.5262, respectively. Additionally, the predicted hallucination rate and Real-Time Factor (RTF) are substantially improved by 96.70\% and 32.38\%, compared to the zero-shot baseline.
\end{abstract}


\begin{highlights}
\item Development of a controlled fine-tuning framework of Whisper-Small with optimized hyperparameters to Automatically Recogninize low-resource Assamese Speech
\item Strategic use of unvalidated crowd-sourced speech data for data augmentation 
\item Establishment of a performance benchmark for Whisper-Small in Assamese speech recognition.
\item Qualitative analysis of orthographic and morphological errors.
\end{highlights}

\maketitle

\section{Introduction} \label{Introduction}

In the age of Artificial Intelligence (AI), Automatic Speech Recognition (ASR) plays a vital role in numerous applications, including automatic video captioning, voice assistants, transcription services, and chatbots \citep{magalhaes2022evaluation}. This makes ASR a crucial bridge to facilitate more effective human-human and human-machine communication \citep{yu2015automatic, sourav2025review}. Out of over 7000 languages globally, ASR systems focus on only a few dominant languages. Currently, ASR research has made remarkable progress in languages with high resources, such as English, Mandarin, Spanish, French, German, Japanese, Persian, Russian, Romanian, and Turkish \citep{leben2018languages,liu2024exploration}. These languages attracted development due to their high economic value \citep{kaur2021automatic}.  Despite breakthroughs, many languages still lag in terms of the availability of high-quality ASR models, especially those with fewer digital resources \citep{javed2022towards}. This disparity highlights the need for focused research and development to build robust ASR systems for low-resource languages (LRLs) such as Assamese, Odia, Bodo, etc. These LRLs lack sufficient linguistic data and resources for effective processing in natural language processing (NLP). Assamese is an Indo-Aryan language having distinct phonological features that make its a speech unique language. It originated from Vedic dialects, and likely developed through Apabhraṃśa from the Magadhi Prakrit of the eastern group of Sanskritic languages. \citep{sarma2012segmentation}. The Assamese language is spoken by a significant population (approximately 15 million) in the northeastern region of India \citep{moral1997north, burling2003northeastern}. Although most Assamese speakers reside in the state of Assam, it is also prevalent in neighbouring North-Eastern states. Some Assamese speakers are also found in Bhutan \citep{mahanta2012assamese}. It is one of the Indo-European languages among the 22 scheduled languages in India and is designated as a ``Classical Indian Language" by the Government of India in October 2024 \citep{sarma2025intersections}.  The Assamese language uses 41 consonants and 11 vowels. There are 20-23 consonant phonemes and 8 distinct vowel phonemes according to the International Phonetic Association (IPA) \citep{mahanta2012assamese}. Acoustic modelling for Assamese ASR systems faces several challenges due to its rich phonemic inventory and the absence of phonemic vowel length distinctions \citep{sarma2014phoneme}. The stress patterns and intonational properties of Assamese further add to the language's complexity. The Assamese language uses overt case marking and a nominative-accusative case system. The language also uses a complex classifier system to categorize nouns by size, shape, animacy, and degree of respect \citep{saikia2019assamese}. Phonetic variations across dialects are another challenge in developing ASR models for Assamese \citep{das2024assamese}. The development of ASR models for Assamese languages remains constrained by the scarcity of annotated speech data and suboptimal model adaptation. Due to its limited representation in publicly accessible speech datasets, the Assamese language, spoken by a sizable population in northeastern India, poses challenges for ASR development. According to recent research, multilingual models such as Whisper and Wav2Vec-BERT can accommodate low-resource languages by using knowledge transfer from high-resource languages \citep{liang2025towards}. Fine-tuning such pre-trained models on language-specific data can further improve their performance, targeting the specific phonetic and linguistic features of the target language \citep{singh2023model}. Another challenge in natural language generation is hallucination \citep{ji2023survey}. Although hallucination is a psychological term often referred to as unreal perception, its definition holds in natural language generation. Characteristics of the Assamese language are prone to several types of hallucination. As Assamese is an agglutinative language, in which suffixes are added to root words to form complex words, it is prone to morphological hallucination. Due to limited resources for finetuning, this often leads to Lexical and Phonetic Fabrications \citep{koudounas2025hallucination}.  

Recent advances in multilingual speech recognition have enabled the development of powerful pretrained models such as Whisper. While these models demonstrate strong performance across many languages, their effectiveness for low-resource languages such as Assamese remains limited due to insufficient language-specific training data and complex linguistic characteristics. Assamese speech presents additional challenges including rich morphology, dialectal variations, and the presence of conjunct characters (Juktaxar), which complicate phoneme-to-grapheme mapping in automatic transcription. Moreover, pretrained multilingual models often generate hallucinated outputs when applied to languages with limited representation in their training data \citep{koenecke2024careless}. These challenges highlight the need for a systematic investigation into controlled fine-tuning strategies that adapt Whisper to Assamese speech while improving transcription reliability and reducing hallucination. Additionally, the availability of a controlled fine-tuning pipeline for Automatic Speech Recognition (ASR) models to transcribe low-resource Assamese language would greatly benefit various applications, including education, healthcare, and government services, by enabling speech-based interfaces and improving content accessibility \citep{javed2022towards}. Motivated by these challenges, this study explores a controlled fine-tuning pipeline for Whisper-based Assamese ASR using publicly available, Mozilla
Common Voice 24.0-Assamese dataset. The key contributions of this study are summarized as follows:

\begin{enumerate}
\renewcommand{\labelenumi}{(\roman{enumi})}
\item Development of a controlled fine-tuning framework with optimized hyper-parameters to mitigate hallucination in Assamese ASR.
\item Strategic use of unvalidated crowd-sourced speech data as a data augmentation resource to improve model robustness.
\item Establishment of a performance benchmark for Whisper-Small in Assamese speech recognition.
\item Qualitative analysis of orthographic and morphological errors, highlighting key challenges in Assamese ASR.
\end{enumerate}

The remainder of this paper is organized as follows. Section \ref{related} provides a review of related works in ASR, with a focus on models and datasets used for low-resource languages, particularly for Assamese. The transformer-based structure of the Whisper model is presented in Section \ref{architecture}. Section \ref{data} presents details on the dataset used in study, including pre-processing, and augmentation techniques used. The evaluation metrics used to assess the model's performance are discussed in Section \ref{evaluation}. Section \ref{method} illustrates the proposed methodology. Experimental results are presented in Section \ref{experiment}. Section \ref{conclusion} concludes and proposes directions for future research. Finally, the limitations of the proposed approach are presented in Section \ref{Limitations}.

\section{Related Works}\label{related}
\begin{figure*}
\centering
\includegraphics[width=0.9\textwidth]{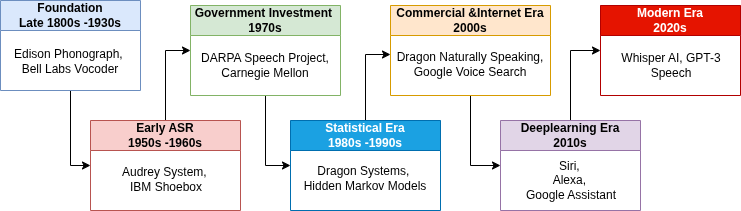}
\caption{Timeline of Automatic Speech Recognition History (1877-2025)}\label{fig_timeline}
\end{figure*}

The foundation for the development of modern Automatic Speech Recognition (ASR) systems began in early 1879, when Thomas Edison introduced his speech-dictation machine. Although this machine could not recognize speech as modern ASR systems do, it was one of the first inventions to record and replay human voices \citep{rabiner2006speech}. The first true ASR, viz. ``Audrey'' was built by Davis et al. of Bell Laboratories in 1952 for the recognition of isolated digits by a single speaker \citep{davis1952automatic}. In the early 1970s, Carnegie Mellon University developed the “Harpy” system as a part of the Speech Understanding Research (SUR) program funded by the Defense Advanced Research Projects Agency (DARPA), U.S.A. This system successfully recognized continuous speech with a vocabulary of 1,011 words and achieved reasonable accuracy at that time \citep{lowerre1990harpy}.  
The field of Automatic Speech Recognition (ASR) has undergone a revolutionary transformation, driven primarily by advances in deep learning technologies. The shift from traditional acoustic models, such as Gaussian Mixture Models (GMMs) and Hidden Markov Models (HMMs), to end-to-end neural network systems has resulted in a significant improvement in transcription quality. The advent of large transformer-based models has established new performance boundaries, as they exhibit an outstanding capacity to generalize across domains and dialects \citep{rabiner1989tutorial}. Figure \ref{fig_timeline} depicts some major milestones in ASR. 

ASR systems for large-resource languages have achieved remarkable success due to the availability of extensive labelled datasets and computational resources \citep{rabiner1989tutorial,baevski2020wav2vec}. These systems are capable of producing near-human transcription accuracy and strong generalization across domains. In contrast, developing ASR systems for low-resource languages poses many challenges, viz., limited annotated speech data, lack of pronunciation lexicons, dialectal variation, and scarce textual resources \citep{radford2023robust}. Limited works have been reported, and it is becoming an active field of research nowadays. Implementation of OpenAI's Whisper model is found in LRLs such as Hakka (Chinese) \citep{chen2023accelerating}, Javanese \citep{timmel2025fine}, Swahili(or Kiswahili) \citep{sharma2025fine}, and Portuguese \citep{perezhohin2024enhancing}. Implementations of the Whisper model for certain Indian low-resource languages (LRLs), such as Rajasthani \citep{bhandari2026post}, and Kannada \citep{prasad2026asr}, have been reported in the literature. However, to the best of our knowledge, no such work has been reported for the Assamese language. Some existing works on Automatic Speech Recognition (ASR) are discussed below.

Sarma et al. \citep{sarma2011assamese} evaluated an Assamese numeral speech recognition system using a hybrid Artificial Neural Network (ANN) that combines Self Organising Maps (SOM) and Multi-Layer Perceptrons (MLP) across diverse recording environments (noiseless, noise mixed, stressed, and stress-free), diverse speaker gender and their emotional states. Medhi and Talukdar \citep{medhi2015isolated} created an isolated-word ASR system for Assamese using 40,000 total utterances and 100 commonly used words, recorded by 20 speakers (10 men and 10 women).  They used k-means to cluster feature vectors, trained an MLP with a single hidden layer, and extracted Zero Crossing Rate, Short-Time Energy, and 12 MFCCs from 25 ms frames.  The system demonstrated the efficacy of combining acoustic features with neural networks for isolated Assamese word recognition, despite difficulties in generalising to new speakers, achieving approximately 99\% accuracy in speaker-dependent tests and approximately 93\% in speaker-independent recognition. Shahnawazuddin et al. \citep{shahnawazuddin2013assamese} design the first deployable moderate-vocabulary Assamese spoken query system for farmers to access the latest agricultural commodity prices via telephone. The integrated ASR model achieved a WER of 15.99\% and 6.13\% for Commodity names and District names, respectively.

Bharali and Kalita \citep{bharali2015comparative} used HTK and HMMs to implement a speaker-independent isolated-word recogniser for Assamese digits 0–9, comparing four acoustic feature sets—LPC, LPC cepstral, mel-filterbank, and MFCC—each with variations such as energy, delta, and acceleration coefficients. The system was tested on data from 15 speakers and achieved up to 80\% accuracy with 39-dimensional MFCC features and a 7-state, 5-hidden-state HMM on clean data, and 95\% accuracy with 26-dimensional LPC cepstral features under noisy conditions. This shows that MFCCs perform best in clean environments, while LPC cepstral features perform best after noise removal.

Agarwalla et al. \citep{agarwalla2016machine} explore machine learning techniques for automatic speech recognition (ASR) using Assamese dialectal speech. The research focuses on using various architectures, including Multi-Layer Perceptron (MLP), Deep Neural Network (DNN), Recurrent Neural Network (RNN), and Fully Focused Time Delay Neural Network (FFTDNN). FFTDNN achieved the highest performance, with Word Recognition Accuracy (WRA) of 95–96\%, outperforming the other architectures. 

Sarma et al.\citep{sarma2017development} built a speech-to-text module with a DNN–HMM hybrid acoustic model. The model achieved 78.05\% word-level accuracy in continuous speech recognition, trained on 3 hours of training data and 0.5 hours of test data. 
Deka et al. \citep{deka2018development} developed a continuous speech recognition (CSR) system for Assamese using a speech corpus of 5,658 utterances from 27 speakers. The proposed DNN-HMM Assamese CSR achieves around 95.7\% word recognition accuracy using scripted speech compared to GMM-HMM and SGMM-HMM (Subspace Gaussian Mixture Models Hidden Markov Models).

Dutta et al.\citep{dutta2022assamese} developed an Assamese speech-based vocabulary identification system using a CNN by collecting 35 isolated words from ten speakers. Using MFCC, spectral centroid, zero-crossing rate, chroma frequencies, spectral roll-off, and intensity features, and comparing CNN against a feed-forward ANN, the CNN achieved 98.4\% accuracy, significantly outperforming the ANN’s 86.4\%.

Various LSTM and BiLSTM-based \citep{kalita2022use} speech recognition systems have been reported, with WER varying between 18.60\% and 18.84\%.

Chen W. et. al.\citep{chen2024data}, fine-tuned an XLSR-53 (cross-lingual speech representations -53) in Marathi, Assamese, and Punjabi from the Common Voice 16.1 corpora. A fine-tuned ASR model achieved the best CER of 19.21\% on the Assamese language. Chen et al. \citep{chen2025owls} also reported a reduction in WER / CER from 65.5 to 29.4 for Assamese using all OWLS models.

\section{Model Architecture} \label{architecture}
\begin{figure*}
\centering
\includegraphics[width=0.7\textwidth]{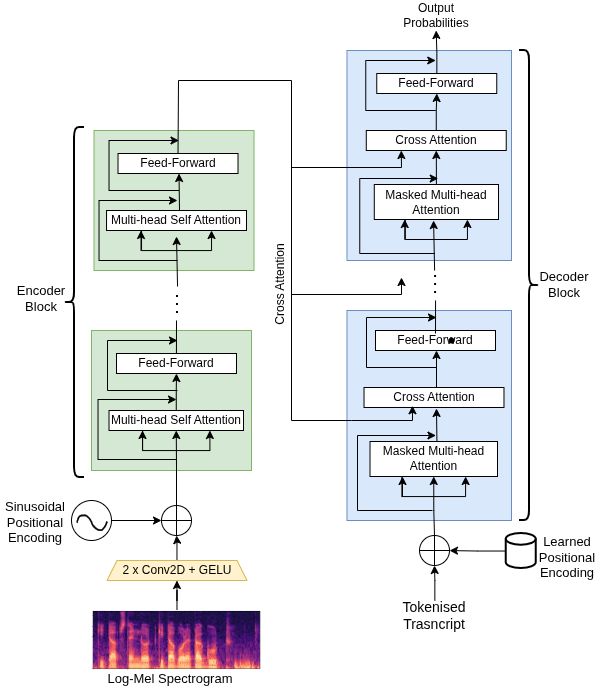}
\caption{Whisper model architecture.}
\label{fig2}
\end{figure*}

OpenAI's Whisper model is a weakly supervised speech processing model that can perform multiple tasks, including ASR, language recognition, speech translation, and speech activity detection, across nearly 100 languages \citep{liu2024exploration}. The capabilities of Whisper on high-resource languages are exceptional, but the same is not true for low-resource languages like Assamese \citep{polat2024implementation, liu2024exploration}. The developer has pre-trained the model using a massive multilingual dataset of 680,000 hours for weakly supervised audio \citep{radford2023robust}.
It has been observed that, for certain languages, Whisper achieves transcription performance that is comparable to, and in some cases exceeds, human-level accuracy under controlled evaluation settings. Additionally, it was found that even a small amount of supervised data can significantly improve the model's word recognition accuracy\citep{liu2024exploration}. As the model is pre-trained on a massive, diverse, multilingual dataset, it provides a robust understanding of phonetics and linguistic patterns that can be generalised for low-resource languages, such as Assamese \citep{de2025whisper}.  The simplified architecture of the Whisper model is shown in Fig.~\ref{fig2}. The model's basic network architecture is a multi-layer stacked Transformer with an encoder-decoder architecture. The model is divided into five versions based on the number of layers, feature dimensions, and attention heads. These versions are named as \texttt{Tiny}, \texttt{Base}, \texttt{Small}, \texttt{Medium}, and \texttt{Large}. Details of each version are presented in the Table \ref{variants}. 


\begin{table}[t]
\caption{Versions of Whisper Family}
\label{variants}
\centering
\renewcommand{\arraystretch}{1.3}
\setlength{\tabcolsep}{2pt} 
\small

\newcolumntype{C}{>{\centering\arraybackslash}X}

\begin{tabularx}{\columnwidth}{lCCCC}
\toprule
\textbf{Model} & 
\textbf{No. of Layers} & 
\textbf{Feature Dimensions} & 
\textbf{Attention Heads} & 
\textbf{Parameters} \\ 
\midrule
Tiny   & 4  & 384  & 6  & 39M  \\
Base   & 6  & 512  & 8  & 74M  \\
Small  & 12 & 768  & 12 & 244M \\
Medium & 24 & 1024 & 16 & 769M \\
Large  & 32 & 1280 & 20 & 1150M \\
\bottomrule
\end{tabularx}
\end{table}
There are several compelling reasons for selecting the Whisper \texttt{`Small'} variant for Assamese ASR. \texttt{`Tiny'} and the \texttt{`Base'} variants are shallow and are incapable of capturing the dental and alveolar sounds of the Assamese language. In our experiment, we observed that the \texttt{`Small'} variant was prone to overfitting and hallucinations. This led to our decision to avoid \texttt{`Medium'} and \texttt{`Large'} variants, as they tend to memorise the training data, leading to overfitting. \texttt{Whisper-small} offer robust generalisation while it is capable of finetuning on consumer-grade GPUs. The \texttt{whisper-small} variant consists of 12 transformer layers each in its encoder and decoder with a hidden size of 768. There are 12 attention heads in both the encoder and decoder. The number of trainable parameters is approximately 244 million.

\section{Dataset and Pre-processing} \label{data}
The efficacy of ASR depends on both the quality and quantity of the data being used \citep{de2014smartphone}. In the following subsection \ref{datasets}, we present details of the dataset used in the study. Preprocessing is performed to feed the data into the Whisper model and to avoid any inconsistency. Details of the preprocessing performed are discussed in subsections \ref{preprocess1} and \ref{preprocess2}.

\subsection{Dataset used} \label{datasets}
The experimental data for this study were sourced from the Common Voice Scripted Speech 24.0 - Assamese \citep{ardila2020common}\citep{mozilla}. The Common Voice Corpus 24.0 is a public, multilingual speech dataset released by Mozilla. The dataset contains 4,681 audio clips, totalling 7.66 hours of speech, of which 3 hours have been formally validated by the community. The dataset is contributed by 51 speakers, of whom 52\% are in their forties. The gender of the contributors was mostly undefined, with 18\% identifying as male. The audio files are in MP3 format, and each is paired with its corresponding text transcription. Along with the validated clips (\texttt{validated.tsv}), it also contains clips that have not been validated by the community (\texttt{other.tsv}). The validated clips were partitioned into official modeling splits: 953 training (\texttt{train.tsv}), 485 development (\texttt{dev.tsv}), and 394 test (\texttt{test.tsv}) samples.  In addition to the official 953 validated training clips (\texttt{train.tsv}), the model was trained on 2,567 unique clips from the `\texttt{other.tsv}' increasing the training set to 3520 clips. 

\subsection{Data Preprocessing} \label{preprocess1}
Data preprocessing is a crucial step in fine-tuning any Acoustic model. Whisper requires both the audio signal and its corresponding transcript as inputs. Therefore, it is essential to normalize these two modalities, and the process of normalization is discussed in the following subsections.

\subsubsection{Audio Signal Normalization}
\begin{figure*}
\centering
\includegraphics[width=.8\textwidth]{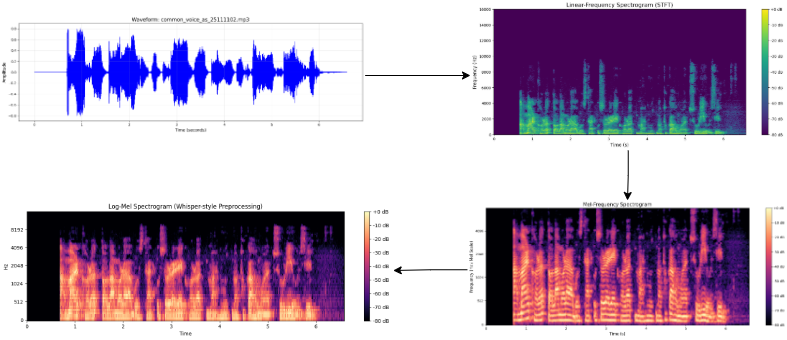}
\caption{Pre-processed Audio required by Whisper model for {\assamese{"আমাৰ ঘৰত তলা মৰা অৱস্থাত ওছৰৰে             ঘৰত মানুহ নথকা সময়ত চুৰি হৈছিল।"}}.(\textipa{[ama\textturnr g\textsuperscript{H}O\textturnr Ot tOla mO\textturnr a ObOxt\textsuperscript{H}at usO\textturnr O\textturnr e g\textsuperscript{H}O\textturnr Ot manuH nOt\textsuperscript{H}Oka xOmOjOt su\textturnr i Hoisil]})}
\label{fig1}
\end{figure*}

A crucial pre-processing step involves preparing the audio files to meet the Whisper model's input requirements. Sampling frequency of the Common Voice audio is 48 kHz. Since most speech processing models, including Whisper, require a sampling rate of 16 kHz, all audio files were resampled to a consistent 16 kHz sampling rate \citep{singh2024comprehensive}. Additionally, stereo sources are converted to a single-channel (\textit{mono}) input. The audio files are then transformed into fixed-size log-mel-spectrogram features, as illustrated in Fig.~\ref{fig1}. The Waveform captures the time-domain utterance and is more informative. However, it is computationally expensive as it contains redundancy \citep{malah1979time}. The first plot is a waveform showing the raw audio amplitude over time. The second plot shows the energy distribution across standard frequency bands, generated by applying the Short-Time Fourier Transform (STFT) to a linear-frequency spectrogram. The Mel spectrogram shown in the third plot represents the converted audio signals in perceptually meaningful Mel scale bands that mimics human auditory perception. It helps in reducing dimensionality while preserving the speaker's perceptual fingerprints. \citep{lambamo2022analyzing}. Finally, the log-Mel spectrogram is generated by applying logarithmic transformation on the Mel scale bands that contains the most informative spectral components required by Whisper\citep{nguyen2025comparative}. Whisper expects audio of a fixed 30-second length. Therefore, the preprocessor pads shorter sentences with silence and truncates longer ones.

\subsubsection{Canonicalization} \label{preprocess2}
In the process of fine-tuning foundation models like Whisper or other ASR models, the quality and consistency of the transcripts are crucial to the model's performance. Canonicalization is a process of converting multiple equivalent textual forms into a single standard representation, is one of the techniques used to achieve such consistency \citep{culotta2007canonicalization}. Assamese language contains several conjuncts ({\assamese{যুক্তাক্ষৰ}}, pronounced as \textipa{[zuk.ta.xO.\textturnr O]}). These conjuncts are formed by combining two or more consonants without an intervening vowel. They can appear in multiple orthographic forms depending on typing or Unicode encoding, making transcript normalization essential.
Additionally, crowd-sourced transcripts of the Common Voice dataset may contain mixed scripts from other languages, such as English. Normalization of the transcripts is performed using NFKC (Normalisation Form Compatibility Composition) Unicode normalisation \citep{davis2001unicode}. NFKC performs aggressive cleaning and standardizes transcripts even for noisy datasets. The advantage of NFKC over other Unicode normalization is that it not only cleans text but also performs orthographic transformations. It defines both canonical and compatibility equivalence to handle complex Assamese scripts. Canonical equivalence concerns a ``visual " match, in which the characters are treated as identical in both appearance and meaning. Characters that have a combination of consonants `{\assamese{ক}}' ([k]) and vowel `{\assamese{ি}}' (\textipa{[i]}) can be represented as `{\assamese{কি}}' (``\textipa{[ki]}'') or ({\assamese{ক + ি}}). NFKC ensures that the Base + mark is collapsed into a single character. Compatibility equivalence addresses ``Functional'' matches, so that characters that might look different due to formatting, such as ligatures, superscripts or subscripts, and stylised variations (bold/italised) represent the same plain-text character. Although there is similarity between the Assamese scripts and their Bengali counterparts, specific glyphs like \textipa{[\textturnr]} ({\assamese{ৰ}}) and \textipa{[wO]} ({\assamese{ৱ}}) are different from Bengali \textipa{[R O]} ({\assamese{র}}) and \textipa{[w]} ({\assamese{ব}}). Because of legacy encoding and the availability of Bengali encoding on most mobile devices, many users may substitute these characters with Bengali characters. Unicode does not consider both Assamese and Bengali `Ra' as equivalent because they belong to different languages. To prevent the model from suffering from Vocabulary Fragmentation, Graphemic Standardization was implemented using custom mapping to unify these representations. 

\subsection{Data Augmentation}
Data augmentation is a common technique to increase the amount of data for training a model\citep{bhat2025two}. \texttt{SpecAugment},  \texttt{Pitch Shifting}, and \texttt{Volume} \texttt{Perturbation} are common data augmentation approaches used in ASR. We employed a data-centric approach for augmentation. The Assamese subset of the Common Voice 24.0 dataset includes two usable subsets, viz., \texttt{validated.tsv} and \texttt{other.tsv} subsets\citep{mozilla}. The validated subset, \texttt{validated.tsv} includes recordings that have received sufficient positive votes from crowd contributors and therefore represent high-quality speech–text pairs. However, the amount of validated audio (approx. 3 hours) is insufficient for fine-tuning Whisper \citep{akera2025much}.  Liu et al. \citep{liu2024exploration}, while fine-tuning Whisper for low-resource African languages, found significant improvements in WER with less than 15 hours of training data. Akera et al. \citep{akera2025much} conducted a data scaling study, demonstrating that practical performance can be achieved with at least 50 hours of transcribed data, with performance continuing to improve up to 200 hours of training data. Considering these findings, the inclusion of additional samples from \texttt{other.tsv} was explored to increase the overall training data available for fine-tuning. The \texttt{other.tsv} subset is comparatively larger, containing nearly three times the \texttt{validated.tsv} set. The \texttt{other.tsv} subset is generally ignored because it lacks crowd-sourced consensus, as only recordings that achieve sufficient positive crowd consensus are included in the \texttt{validated.tsv} set. This could be due to dialectical differences or the presence of background noise. This anomaly can serve as a natural regularisation mechanism for fine-tuning the ASR model. The availability of this additional data (\texttt{other.tsv}) enables implementing the data-centric augmentation approach alongside other augmentation techniques, such as \texttt{SpecAugment}. We prepared the training corpus by merging the train subset (\texttt{train.tsv}) with \texttt{other.tsv}. Care has been taken to maintain the integrity of the test  (\texttt{test.tsv}) and validation (\texttt{dev.tsv}) sets, removing them from the merged data so that the model does not accidentally train on these data. 

\section{Evaluation Metrics Used} \label{evaluation}
Different evaluation metrics are available to measure the efficiency of an ASR Model. The primary metrics used in this study are, viz.,  Word Error Rate (\textit{WER}), Character Error Rate (\textit{CER}), Match Error Rate (\textit{MER}), BLEU Score (Bilingual Evaluation Understudy), METEOR (Metric for Evaluation of Translation with Explicit ORdering), Word Information Lost (\textit{WIL}), \textit{Precision}, \textit{Recall}, \textit{F1-score}, Real-Time Factor (\textit{RTF}). These metrics are discussed in this section.

\subsection{Word Error Rate (WER)}
Word Error Rate (WER) is a widely adopted evaluation metric that measures the accuracy of a predicted transcript relative to a ground-truth reference \citep{morris2004and}. WER is computed as 
\begin{equation}
WER = \frac{\Sigma{(\mathcal{S}, \mathcal{D},\mathcal{I}})}{\mathcal{N}},
\end{equation}
where $\mathcal{S}$, $\mathcal{D}$, and $\mathcal{I}$ correspond to substitution, deletion, and insertion errors, respectively, and $N$ denotes the total number of words in the reference transcription. Substitution errors occur when an incorrect word replaces a reference word, deletion errors arise when a reference word is omitted, and insertion errors correspond to additional words that do not appear in the reference transcription. A lower WER indicates better recognition performance. Metrics like Word Recognition Rate (WRR) can be computed from WER as $WRR = 1-WER$. 

\subsection{Character Error Rate (CER)}
The Character Error Rate (CER) is another commonly used evaluation metric for assessing the accuracy of an ASR system. It provides a fine-grained evaluation of transcription quality by measuring character-level recognition errors rather than word-level errors\citep{morris2004and}.
\begin{equation}
CER =\frac{\Sigma{(\mathcal{S}, \mathcal{D},\mathcal{I}})}{\mathcal{N}}
\end{equation}
\indent Where $\mathcal{S}$,$\mathcal{D}$, $\mathcal{I}$, and $\mathcal{C}$ are the total number of substitutions, deletions, insertions, and correct words, respectively. $\mathcal{N} =\Sigma{(\mathcal{S}, \mathcal{D},\mathcal{C}})$ is the total number of characters in the reference transcript. CER is useful for languages with agglutinative morphology and complex word boundaries.
Both metrics provide a quantitative assessment of the model's transcription accuracy.
\subsection{Match Error Rate (MER)}
Match Error Rate ($MER$) measures the word-level errors relative to the total number of aligned words between the reference and predicted transcripts\citep{morris2004and}. 
\begin{equation}
MER = \frac{\Sigma{(\mathcal{S}, \mathcal{D},\mathcal{I}})}{\Sigma{(\mathcal{C},\mathcal{S}, \mathcal{D},\mathcal{I}})}
\end{equation}

\indent where $\mathcal{C}$ denotes the number of correctly recognized words, $\mathcal{S}$ represents the number of substitutions, $\mathcal{D}$ denotes the number of deletions, 
and $\mathcal{I}$ denotes the number of insertions in the predicted transcription.

\subsection{Word Information Lost (WIL)}
The Word Information Lost (WIL) metric measures numerically how much word-level information is lost between the reference and the predicted transcription by counting the substitutions, deletions, and insertions of the words,  providing a more information-theoretic assessment than the simple Word Error Rate (WER)\citep{morris2004and}. Mathematically, WIL is defined as 

\begin{equation}
\text{WIL} = 1 - \frac{\mathcal{C}}{\mathcal{N}} \cdot \frac{\mathcal{C}}{\Sigma{(\mathcal{C}, \mathcal{S},\mathcal{I}})}
\end{equation}

\indent where $\mathcal{C}$ is the number of correctly recognized words, $\mathcal{S}$ the number of word substitutions, $\mathcal{I}$ is number of word insertions, and $\mathcal{N}$ is the total number of words in the reference. A Lower WIL value indicates better predicted transcription accuracy. WIL combines correctness relative to both the reference and predicted transcripts.

\subsection{BLEU Score (Bilingual Evaluation Understudy)}
The BLEU Score is a corpus-based automatic evaluation measure widely used for machine translation, Automatic Speech Recognition (ASR)  \citep{papineni2002bleu}. The BLEU score is language independent and closely resembles human judgment. It analyze the similarity between a generated hypothesis and the reference texts using \textit{modified n-gram precision} ($P_n$), while penalizing excessively short predicted candidate sentences through a \textit{brevity penalty}. Higher BLEU scores indicate stronger correspondence to the reference transcript. Calculation of \textit{modified n-gram precision} ($P_n$), \textit{brevity penalty}, and Overall BLUE score is given below. 

\textbf{Modified n-gram precision($P_n$):} Modified n-gram precision value ($P_n$) is the measure of the proportion of n-grams in a candidate sentence that correctly appear in one or more reference sentences. In standard n-gram precision, every occurrence of an n-gram in the candidate is counted, even if it appears more times than in the reference, while in the modified n-gram precision, the count of each n-gram in the candidate is clipped to the maximum number of times it appears in the reference. This ensures no  of n-gram computation. $P_n$ is calculated using the following formula
\begin{equation}
P_n = \frac{\sum_{g \in C} \min\big(\text{Count}_C(g), \text{Count}_R(g)\big)}{\sum_{g \in C} \text{Count}_C(g)}
\end{equation}

where, $C$ represents candidate sentence, $R$ denotes reference sentence(s), $g$ is the n-gram of length $n$, $\text{Count}_C(g)$ represents the frequency of n-gram $g$ in the candidate, and $\text{Count}_R(g)$ denotes the frequency of n-gram $g$ in the reference. The expression, $\min\big(\text{Count}_C(g), \text{Count}_R(g)\big)$ ensures that no n-gram is over-counted. 

\textbf{Brevity penalty (BP):} \textit{Brevity penalty (BP)} is calculated to prevent artificially high precision cores excessively caused by shorter predictions than the reference, which could be achieved by omitting the words present in the reference but missing in the candidate. The mathematical expression for $BP$ is 
\begin{equation}
BP = 
\begin{cases} 
1 & \text{if } c > r \\[1ex]
\exp(1 - r/c) & \text{if } c \le r
\end{cases}
\end{equation}
\indent where $c$ and $r$ are the lengths of the candidate and reference, respectively.

\textbf{Overall BLEU score:} Finally, the overall BLEU score is calculated that combines multiple n-gram precisions using a geometric mean, scaling by the Brevity penalty using the following mathematical formula:
\begin{equation}
\text{BLEU} = BP \cdot \exp \left( \sum_{n=1}^{N} w_n \log P_n \right)
\end{equation}
\indent where $N$ is the maximum n-gram order (commonly 4) and $w_n$ are the weighting coefficients, typically uniform ($w_n = 1/N$).

This formulation ensures that BLEU evaluates both lexical overlap and fluency at the n-gram level, while discouraging artificially short candidate outputs.

\subsection{METEOR (Metric for Evaluation of Translation with Explicit ORdering)}
 METEOR is more flexible than BLEU, offering matching techniques such as exact, stem, synonym, and paraphrase matches. 
 METEOR is frequently regarded as a more accurate reflection of human quality assessment\citep{banerjee2005meteor}. METEOR computes unigram precision (P) and recall (R) by counting the number of correctly matched words in the reference and predicted transcriptions. Based on the precision ($P$) and recall ($R$), a weighted harmonic mean, $\mathbb{F}_{mean}$ is computed\citep{labied2024assessing, magalhaes2022evaluation}.
 
 \begin{equation}
      \mathbb{F}_{mean} = \frac{10PR}{R + 9P}
 \end{equation}

 METEOR introduces a fragmented penalty factor ${M}_{penalty}$, which is used to reduce the METEOR score even when the matched words between the reference and predicted are scattered, i.e., they appear in fragmented segments rather than a continuous sequence.
 \begin{equation}
    {M}_{penalty} = \alpha \left(\frac{ch}{m_n}\right)^{\beta}
\end{equation}
 
\indent Where, $ch$ reperent the number of chunks (continuous matched sequences), $m_n$ denotes the number of matched unigrams, $\alpha$ denotes the penalty parameter, and $\beta$ denotes the fragmentation parameter.

The final METEOR score is calculated by multiplying $\mathbb{F}_{mean}$ by a penalty factor, ${M}_{penalty}$. The mathematical expression for the final METEOR calculation is as follows 

\begin{equation}
{M}_{final} = \mathbb{F}_{mean} \times (1 - {M}_{penalty})
\end{equation}

\subsection{Precision}
In machine learning, Precision metric evaluates the proportion of true positive predictions out of all positive predictions made by a given model. In case of ASR, True positive predictions correspond to the correctly recognized words in the predicted transcripts with respect to the total number of words generated by the predicted transcripts\citep{kapusta2024text}.
Mathematically, precision is defined as
\begin{equation}
Precision = \frac{TP}{TP + FP}=\frac{\mathcal{C}}{\mathcal{C} + \mathcal{I}}
\end{equation}

\indent Where, $TP$ (True Positives) represent correctly predicted words($\mathcal{C}$), and $FP$ (False Positives) represent incorrectly predicted tokens that do not exist (i.e. Insertion, $\mathcal{I}$) in the reference transcript. A higher precision value of an ASR indicates fewer incorrect insertions in the predicted transcripts.

\subsection{Recall}

Recall of an ASR model measures the proportion of words in the reference transcripts that are correctly recognized in the predicted is transcript\citep{morris2004and}. Mathematically, Recall is defined as

\begin{equation}
Recall = \frac{TP}{TP + FN}=\frac{\mathcal{C}}{\mathcal{C} + \mathcal{D}}
\end{equation}

\indent Where, $TP$ (True Positives) denotes the correctly predicted words ($\mathcal{C}$), and $FN$ (False Negatives) denotes the number of words in the reference transcript that were missed (i.e. deletion, $\mathcal{C}$) by the ASR model.
\subsection{F1-score}
In ASR evaluation, the \textit{F1-score} measures the balance between the accuracy of predicted words and the coverage of reference words. It is defined as the harmonic mean of Precision and Recall\citep{morris2004and, el2021evaluation}:
\begin{equation}
\begin{aligned}
\textit{F1-score}
&= 2 \cdot \frac{\text{Precision} \cdot \text{Recall}}
{\text{Precision} + \text{Recall}} \\
&= \frac{2\,TP}{2\,TP + FP + FN} \\
&= \frac{2\,\mathcal{C}}
{2\,\mathcal{C} + \mathcal{I} + \mathcal{D}}
\end{aligned}
\end{equation}

\textit{F1-score} helps to evaluate overall word-level accuracy.

\subsection{Hallucination Error Rate(HER)}
In Automatic Speech Recognition (ASR), hallucination occurs when the system produces tokens that do not exist in the audio signal.
Hence, the Hallucination Error Rate (HER) quantifies how often a model generates output that is not supported by the input

Let the reference transcript be defined as

\[
\text{REF} = (r_1, r_2, \ldots, r_n), 1 \le n\le 
\mathcal{N}
\]
\indent where $\mathcal{N} = |\text{REF}|$ denotes the number of tokens in the reference transcript. \\
Let the corresponding predicted transcript of above $REF$ generated by the ASR system be

\[
\text{PRED} = (p_1, p_2, \dots, p_m), 1 \le m\le 
\mathcal{M}
\]
\indent where $\mathcal{M} = |\text{PRED}|$ represents the number of predicted tokens.

An alignment function $\mathcal{A}$ maps each predicted token to either a corresponding reference token or to $\varnothing$, if no valid alignment exists, and it can be defined as follows
\[
\mathcal{A} : \{1,\dots,\mathcal{M}\} \rightarrow \{1,\dots,N\} \cup \{\varnothing\}
\]

The alignment, $\mathcal{A}$ is typically obtained using minimum edit distance i.e. \textit{Levenshtein alignment}.

The set of hallucinated tokens is defined as
\[ \mathcal{H} = \{\, p_i \in \text{PRED }  | \mathcal{A}(i) = \varnothing \,\}, 1 \le i \le \mathcal{M} \]
\indent Where, $\mathcal{A}(i) = \varnothing$, represents $i$-th tokens in the prediction that do not align with any token in the reference transcript.

The Hallucination Error Rate ($HER_{ref}$) normalized with respect to the reference transcript is therefore given by the following equation

\begin{equation}
HER_{ref} =
\frac{|\mathcal{H}|}{|\text{REF}|}
=
\frac{
\left|
\{\, p_i \in \text{PRED} : \mathcal{A}(i)=\varnothing \}
\right|
}{\mathcal{N}}.
\end{equation}

However, normalizing hallucinations with respect to the reference length may inflate the hallucination estimation, particularly when the model generates longer transcripts than the reference transcript. To mitigate this bias, $HER_{pred}$ can alternatively be defined relative to the predicted transcript length as

\begin{equation}
HER_{pred} =
\frac{|\mathcal{H}|}{|\text{PRED}|}
=
\frac{
\left|
\{\, p_i \in \text{PRED} : \mathcal{A}(i)=\varnothing \}
\right|
}{\mathcal{M}}.
\end{equation}

\subsection{Real-Time Factor (RTF)}

Real-Time Factor (\textit{RTF}) is a metric used to evaluate the computational efficiency of an ASR system. It is defined as the ratio of the total processing time, ${T_p}$, taken by the ASR system to process an input audio signal to the duration,  $T_d$, of the audio itself.

\begin{equation}
\textit{RTF} = \frac{T_p}{T_d}
\end{equation}

An \textit{RTF} $\leq 1$ indicates that the ASR system can process speech faster than real time, whereas \textit{RTF} $> 1$ indicates slower processing than real time.

\section{Methodology} \label{method}
This research aims to develop a robust ASR for the low-resource Assamese language. 
The process begins with selecting the appropriate foundation model. Our methodology for fine-tuning Wishper is organized in four distinct stages, viz., \texttt{Zero-Shot Evaluation}, \texttt{Controlled Fine-tuning approach}, \texttt{Optimization} and \texttt{Decoding Strategy}.
Zero-shot evaluation is performed to establish a benchmark, followed by controlled fine-tuning to mitigate hallucinations. To maximize the model's performance on Assamese speech recognition, we implemented a robust optimization framework alongside specialized decoding strategies during the fine-tuning and inference stages. Details of all these stages are discussed in the following sections.

\subsection{Zero-Shot Evaluation} \label{zeroshot}
Zero-shot evaluation involves assessing a machine learning model on an unfamiliar task, dataset, or language without any targeted training or fine-tuning. Zero-Shot evaluation of the Whisper-small model was performed by testing the multilingual pre-trained model on Assamese speech data (\texttt{test.tsv}) without task-specific fine-tuning. This was conducted to evaluate how well Whisper generalizes to Assamese. Additionally, an evaluation was also conducted to establish a benchmark for assessing the fine-tuned model's performance. Let $\mathcal{M}_\theta$ denote the transformer-based Whisper-small model parameterized by $\theta$. The model is pre-trained on a massive, heterogeneous dataset $\mathcal{D}_{train}$, so that it assigns high probability to the correct sequence of words $Y = \{y_1, y_2, \dots, y_n\}$ conditioned on the acoustic feature set $X$ 
\begin{equation}
    \mathcal{L}(\theta) = -\mathbb{E}_{(X, Y) \sim \mathcal{D}_{train}} \left[ \sum_{i=1}^{n} \log P(y_i \mid y_{<i}, X; \theta) \right]
\end{equation}
\indent where, \\
\indent $\mathcal{L}(\theta)$ is the Loss function, $\mathbb{E}_{(X, Y) \sim \mathcal{D}_{train}}$ is the expectation operator across the entire pre-training dataset $\mathcal{D}_{train}$, $y_{<i}=\{y_1, y_2, \dots, y_{(i-1)}\}$ is the sequence of previous tokens, and $P(y_i \mid y_{<i}, X; \theta)$ is probability that the model $\mathcal{M}_\theta$ assigns to the correct token $y_i$ for input audio $X$ for parameters $\theta$

The zero-shot evaluation is performed using a target domain $\mathcal{D}_{target}$ (\texttt{test.tsv} from Common Voice 24.0 Assamese speech), under the assumption of data isolation where $\mathcal{D}_{target} \cap \mathcal{D}_{train} = \varnothing$. The model generates the most probable sequence $\hat{Y}$ via beam search while the parameter set $\theta$ remains unchanged.

\begin{equation}
    \hat{Y} = \arg\max_{Y \in \mathcal{Y}^*} \prod_{i=1}^{n} P(y_i \mid y_{<i}, X; \theta)
\end{equation}

The discrepancy between the predicted sequence $\hat{Y}$ and the ground-truth $Y_{ref} \in \mathcal{D}_{target}$ is quantified using the WER, CER, MER, BLEU Score, METEOR, WIL, Precision, Recall, F1-score, HER and RTF.

\begin{figure*}
\centering
\includegraphics[width=.9\textwidth]{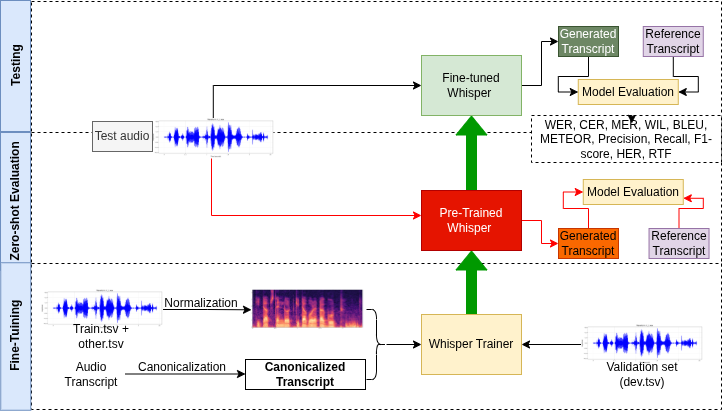}
\caption{Zero-shot, Fine-tuning, and Testing Methodology.}
\label{fig3}
\end{figure*}

\subsection{Controlled Fine-Tuning}
Controlled Fine-tuning of the Whisper model was performed using the Hugging Face Transformers library version 5.2.0, running on Python version 3.12.12. This library provides a Trainer class designed for efficient model training. Fine-tuning the pre-trained Whisper model was performed to adapt it to the unique lexical and phonetic characteristics of the Assamese language. The controlled fine-tuning and inference pipeline used in this study is illustrated in Figure 4. The process begins with audio normalization and canonicalization of the transcripts, ensuring consistent and standardized input for the Whisper Trainer. The fine-tuned Whisper model is then used to generate transcripts for the audio in the \texttt{test.tsv} file. The predicted transcripts are then compared against the reference transcripts to evaluate performance using Word Error Rate (WER) and Character Error Rate (CER), Match Error Rate (MER), Word Information Lost (WIL), BLEU Score (Bilingual Evaluation Understudy), METEOR,  Precision, Recall, F1-score, Hallucination Error Rate(HER), and Real-Time Factor (RTF). The results provide the quantitative measures of transcription accuracy of the fine-tuned model performance at both the word and character levels.

In standard fine-tuning, the dataset is typically assumed to contain perfect ground truth annotations. However, it is observed that the available validated data in Assamese language is not sufficient. To encounter this limitation, we incorporated the lower-consensus audio data (\texttt{other.tsv}) subset alongside the validated set during fine-tuning. These data may contain acoustic noise. Hence, specific measures through Unicode NFKC normalization is required to sanitise the transcripts preventing the model from learning noise patterns instead of phonetic characteristics. Incorporation of \texttt{other.tsv} introduced acoustic variations, but special care had to be taken to ensure that any non-Assamese character inclusion is normalised. All punctuations were also removed, to allow the model to concentrate on exclusively learning the phonetics, rather than wasting the time of learning grammatical pauses. 
Fine-tuning of the model on insufficient Assamese data may led to overfitting. This requires regularizing the model by applying a linear learning rate warmup (LRW) and weight decay. 
The fine-tuning process is further supported by a high-consensus validation set, \texttt{dev.tsv}, which guided model optimisation and early stopping.
 
\subsection{Model Optimization}
As Whisper-small is pre-trained on large-scale multilingual speech data containing 244 million parameters, an efficient and scalable deployment of the model requires certain optimizations to reduce computational load while maintaining transcription accuracy and robustness. Both hardware-level and model-level optimizations were performed to improve resource utilization while maintaining the model's robustness. These optimizations are discussed in the following sections.

\subsubsection{Hardware-Level Optimization}
Experiments were conducted in a Kaggle T4 GPU environment with a VRAM limit of 16GB. This available GPU memory presents a constraint when fine-tuning large models like Whisper. Because of this constraint, several hardware-related optimization techniques were employed. 
We kept the training batch size at 2 with 4 Gradient Accumulation Steps. This allowed the model to effectively learn from 8 samples before updating its weights. The Whisper-small variant with 244M parameters often encountered an ``Out of Memory" error with larger batch sizes. Standard fine-tuning uses 32-bit floats (FP32), but due to VRAM constraints, we employed mixed-precision with 16-bit floats (FP16).

\subsubsection{Model-level Optimizations}
The model-level optimization was performed using linear learning rate warmup (LRW) and Weight decay. LRW gradually increases the learning rate from a small value to the target learning rate during the warmup steps of training \citep{goyal2017accurate}. Given the targeted learning rate $\eta$, learning rate $\eta_{t} $, at any given step $t$ is a fraction of the targeted learning rate, calculated as 

\begin{equation}
\eta_{t} = \eta \times \frac{t}{t_{w}}, t \leq t_{w}
\end{equation}

where $t_{w}$ is the total number of steps in the warmup phase. A linear learning rate warm-up was employed for the initial 50 steps. This helped in the gradual increase in the learning rate before reaching its peak at $1\times 10^{−5}$. This scheduling strategy facilitates the acquisition of initial statistics to support reliable updates. Furthermore, LRW also helped in preserving the multilingual knowledge acquired during pre-training, as it is exposed to a new target language. As the Zero-shot evaluation revealed, the model had limited exposure to Assamese during pre-training, which could cause its gradients to be unstable at first when it encounters Assamese speech. LRW also controlled this volatility. To facilitate stable convergence, the optimization process was governed by a cosine annealing learning rate scheduler that gradually decreases the learning rate to near zero as training progresses.

\begin{equation}
    \eta_t = \eta_{min} + \frac{1}{2}(\eta_{max} - \eta_{min}) \left( 1 + \cos\left( \frac{t}{T_{max}} \pi \right) \right)
\end{equation}
where, $T_{max}$ represents the total number of training steps and $\eta_{min}$ is the terminal learning rate.

Weight decay introduced by Hanson and Pratt \citep{hanson1988comparing} is a regularization technique that adds a penalty to the loss function based on the size of the weights. It is equivalent to the L2 regularization, where penalty is added to $\mathcal{L}({\theta})$, 

\begin{equation}
\mathcal{L}_{\text{total}}(\theta) = \mathcal{L}(\theta) + \frac{\lambda}{2}||\theta||_2^2
\end{equation}

where $\theta$ is a vector containing all weights (\textit{w}) and biases (\textit{b}), $\lambda$ is the Weight Decay Coefficient, and $||\theta||_2^2$ denotes the squared L2 norm calculated as $ ||\theta||_2^2 = \sum_{j=1}^{n} \theta_j^2$. At each training step $t$, the weights $\theta$ are updated as follows 
\begin{equation}
\theta_{t+1} = (1 - \eta \lambda)\theta_t - \eta \nabla L(\theta_t) 
\end{equation}

To mitigate model overfitting, we used a weight decay of 0.05 in our approach. With 244M parameters, the model is prone to memorization with merely a few thousand training samples. A large weight decay of 0.05 adds a penalty to the loss function based on the magnitude of the weights, forcing the model to find the simplest mathematical path to the correct transcription. The default AdamW (Adaptive Moment Estimation with Decoupled Weight Decay) optimizer was employed to update the model parameters  \citep{loshchilov2017decoupled}.  AdamW is based on Adam, which decouples the weight decay regularization from the gradient update step. The steps of adaptive gradient and the decoupled weight decay used in AdamW are presented in Equation \ref{adamW}. 
\begin{equation} \label{adamW}
\theta_{t+1} =
\underbrace{\theta_t - \eta \frac{\hat{m}_t}{\sqrt{\hat{v}_t} + \epsilon}}_{\text{Adaptive Gradient Update}}
-
\underbrace{\eta \lambda \theta_t}_{\text{Decoupled Weight Decay}}
\end{equation}

where $\theta$ is a weight vector $\eta $ is the Learning Rate, $\lambda$ is the Weight Decay coefficient, $\hat{m}_t$ is the momentum, $\hat{v}_t$ is the uncentered variance and $\epsilon$ is a small constant to prevent division by zero.
AdamW applies weight decay directly to the parameters rather than through the gradient-based moments. Use of AdamW enables precise, parameter-specific updates for the Assamese phonemes, which contain several conjuncts.  We incorporated a light dropout rate of 0.05 into the transformer attention and fully connected layers to enhance the model's robustness. Stochastically deactivating a subset of neurons prevented the development of dependencies between units. The value of 0.05 was strategically selected to preserve the pre-trained knowledge while mitigating the architectural redundancy. The model's performance was evaluated every 500 steps against the validation set. Early Stopping with a patience of 2 (two) evaluation cycles (1000 steps) terminates fine-tuning when validation loss stops improving, even if the training loss continues to decrease.  The hyperparameters configured for the training session are presented in 
Table~\ref{hyperparameter}. 

\begin{table}
\caption{Hyperparameters used in fine-tuning}
\label{hyperparameter}
\centering
\begin{tabular}{lll}
\toprule
\textbf{Category} & \textbf{Hyperparameter} & \textbf{Value} \\
\midrule

\multirow{3}{*}{Optimization}
 & Learning Rate & $1 \times 10^{-5}$ \\
 & Warmup Steps & 50 \\
 & Optimizer & AdamW \\

\midrule
\multirow{3}{*}{Efficiency}
 & Precision & FP16 \\
 & Grad. Accumulation & 4 \\
 & Grad. Checkpointing & Enabled \\

\midrule
\multirow{3}{*}{Regularization}
 & Weight Decay & 0.05 \\
 & Dropout Rate & 0.05 \\
 & Early Stopping & 2 (patience) \\

\midrule
\multirow{2}{*}{Training Flow}
 & Max Steps & 3000 \\
 & Evaluation Freq. & 500 Steps \\

\bottomrule
\end{tabular}
\end{table}

\subsection{Decoding Strategy}
In the Decoder block of Whisper, the determination of the next Assamese character depends on the search strategy employed. There is a trade-off between speed and linguistic farsightedness while deciding on the search strategy. A greedy search at each timestamp decides the next token by selecting the one with the highest absolute probability, ignoring remaining tokens. This search strategy is simple and fast, but may not be sufficient in handling agglutinative languages like Assamese. To mitigate this problem, the Beam Search  \citep{lowerre1990harpy, greer1982acoustic} approach is chosen with a beam width of 5. In contrast to greedy search, beam search tracks the 5 most likely sequences concurrently. Subsequently, the cumulative probabilities of these 25 paths are calculated, and the top 5 are maintained. Thus, the Beam search acts as a probabilistic filter to reduce hallucinations.

\section{Experimental Results} \label{experiment}
Experimental evaluation of controlled fine-tuning of \texttt{Whisper-small} demonstrates a transformative shift in performance from a hallucinating pre-trained multilingual model to a robust ASR. The experimental results of the fine-tuned Whisper-small model, both quantitative and qualitative, are discussed in the following sections.

\subsection{Quantitative Analysis} \label{quantitative}
Quantitative analysis of an ASR model involves measuring model performance using numerical metrics, such as Word Error Rate (WER) and Character Error Rate (CER), providing an objective evaluation of overall transcription accuracy. The Fine-tuning log of the Whisper model provides important information about its behaviour during the tuning phase. These logs are analyzed in the following section. A comparison of the model is also presented before and after fine-tuning in the subsequent section. Finally, the reasons for difference in the evaluation metrics are also investigated.

\subsubsection{Zero-shot Evaluation Results (Baseline)}
In our study, zero-shot evaluation provides a benchmark demonstrating that \texttt{Whisper-small} lacks the ability to correctly transcribe Assamese. For the zero-shot evaluation, the pre-trained Whisper-small model was tested using the test data of the Common Voice 24.0 Assamese subset. The findings indicate poor transcription performance in the zero-shot condition, evidenced by a high Word Error Rate (WER) of 2.0127 and a Character Error Rate (CER) of 1.9091, suggesting significant discrepancies between the predicted and actual transcriptions. Likewise, the Match Error Rate (MER) and Word Information Lost (WIL) reached 1.0000, indicating a serious decline in both recognition accuracy and linguistic alignment. Additionally, text quality measures reinforce this shortcoming, with both BLEU and METEOR scores recorded at 0.00, highlighting the model's failure to produce outputs that are either semantically or structurally coherent in the unfamiliar domain. The Precision, Recall, and F1 scores remained zero, indicating that accurate token-level predictions were very limited. Additionally, the hallucination analysis revealed high values, with HER$_{ref}$ at 1.0127 and HER$_{pred}$ at 0.5552, which suggests frequent occurrence of generating unsupported or misleading tokens. InDespite these performance challenges, the real-time factor (RTF) of 0.2943 indicates a swift inference speed, demonstrating that computational efficiency is maintainedven as recognition performance declines. On further examination of some random samples, it is found that the model encountered a language it did not know well, producing repetitive punctuation (, , ,  \dots) or single characters (d d d d d \dots) in some predicted transcripts. In other predicted transcripts, it identified the word phonetically but represented the transcripts in Hindi, Arabic, and some in English. The process of producing such kind of output is termed as hallucination \citep{ji2023survey}. To be more specific, the model is suffering from cross-lingual hallucination. Examples of some of these transcripts produced by Whisper-small are presented in Table \ref{samplepredzero}. Zero-shot with Whisper-medium also produced similar results. The knowledge transfer gap is clearly visible from the results. A worst-case scenario of hallucination is also seen. 
Overall, the zero-shot results establish a challenging baseline and emphasize the necessity of domain-specific fine-tuning to achieve reliable transcription performance in low-resource or previously unseen linguistic settings.

\begin{table*}
\caption{Zero-shot Prediction Samples of Whisper-small}
\label{samplepredzero}
\centering
\begin{tabular}{p{2.5cm} p{13cm}}
\toprule
\textbf{Sample ID} & \textbf{Reference and Prediction} \\
\midrule
Sample 5 & REF: {\assamese{কোৱা হয় সেই পীঠৰ মন্দিৰ সৰ্ব্ব প্ৰথম নিৰ্মাণ কৰিছিল নৰকাসুৰে}}\\ 
& PRED: {ḍ្្្្្្្្្្្្្្្្្្្្្្្្្្្្្្្្្្្្្្្្្្្្្្្្្្្្្្្្្្្្្្្្្្្្្្្្្្្្្្្្្្្្្្្្្្្្្្្្្្្្្្្្្្្្្្្្្្្្្្្្្្្្្្្្្្្្្្្្្្្្្្្្្្្}\\
\midrule

Sample 12 & REF: {\assamese{কন্দা দেখি মাকৰ কোলাত হেৰি উঠিলগৈ}}\\
& PRED: {\hindi{खंदा देखी माकर्खला देडी उतिल कोई }}\\
\midrule

Sample 41 & REF: {\assamese{এবছৰ আগেয়ে কানাডালৈ পলাই গৈছিল}}\\
& PRED: {\hindi{आब श्वर अगे एक खनादाल। पोलाये गुइसिल}}\\
\midrule

Sample 52 & REF: {\assamese{এবছৰ আগেয়ে কানাডালৈ পলাই গৈছিল}}\\
& PRED: {\hindi{आब श्वर अगे एक खनादाल। पोलाये गुइसिल}}\\
\midrule

Sample 57 & REF: {\assamese{ঘৰৰ গৃহিণীৰ ওপৰতে গোটেই পৰিয়ালৰ সুখ দুখ নিৰ্ভৰ কৰে}}\\ 
& PRED: {unknown repeated character}\\
\midrule

Sample 71 & REF: {\assamese{মণ্ডলৰ ঘৰৰ পৰা খবৰ আহিল দৰা এতিয়াও ভাল হোৱা নাই}}\\ 
& PRED: {món dolor kóror bóra kóbor ahi, dorá iti áo pál ho anái.}\\
\midrule

Sample 75 & REF: {\assamese{সি চাগে তোমাৰ কথা শুনা নাই আগতে}}\\ 
& PRED: {\hindi{इसागे तुमार कोटा खुनाना यागोते}}\\
\midrule

Sample 114 & REF: {\assamese{আগতে বন্দবস্ত হৈছিল আপোনালোকৰ ফালৰ মদন মণ্ডলৰ ভায়েকৰ লগত}}\\ 
& PRED: {\hindi{अगो ते बन्दबश्त हो शिल आपना लुकर फलर मदन मन्दलर भायागर दगत}}\\
\midrule

Sample 140 & REF: {\assamese{কাণ্ডকাৰখানা দেখি লভিতা আৰু লগৰ ছোৱালী কেজনী নিজৰ ঘৰৰ দুৱাৰমুখত থুপ খালেগৈ}}\\
& PRED: {\hindi{तो तो तो तो तो तो तो तो तो तो तो तो तो}}$\dots$\\
\midrule

Sample 182 & REF: {\assamese{লভিতাহঁতে উত্তেজিত হৈ বাহিৰৰ ফালে চাবলৈ ধৰে}}\\
& PRED: {\hindi{लविता हाते उटे जितो है बाहिर अर्फाले सबले रहारे}}\\
\midrule

Sample 276 & REF: {\assamese{বিশেষকৈ শ্ৰীমান অভয় দুৱৰা আৰু শ্ৰীমান তৰুণ দুৱৰাৰ বাবেইহে}}\\ 
& PRED: {ḍḍḍḍḍḍḍḍḍḍḍḍḍḍ}$\dots$\\
\midrule

Sample 327 & REF: {\assamese{কোৰআন শ্বৰীফ ইছলাম ধৰ্মৰ মূল ধৰ্মগ্ৰন্থ  }}\\
& PRED: {\hindi{तगुराम् सुरिप इस्लाम्द्रमार मुल्द्रमागान्ता}}\\
\midrule

Sample 345 & REF: {\assamese{মানুহটোৱে কলে ভালকৈ শিক নিশিকিবি}}\\
& PRED: {\hindi{मनु ट्ट्ट्ट्ट्ट्ट्ट्ट्ट्ट्ट्ट्ट्ट्ट्ट्ट्ट्ट्ट्ट्ट्ट्ट्ट}}\\
\midrule

Sample 377 & REF: {\assamese{তহঁতৰ নো খায় কোনে}}\\ 
& PRED: {\hindi{तो हतर तो खाय कुने}}\\
\midrule

Sample 379 & REF: {\assamese{মাষ্টাৰ গলেই হয়}}\\
& PRED: {\hindi{वाँदादादादादादादादादादादादादादादा}} $\dots$\\

\bottomrule
\end{tabular}
\end{table*}

\subsubsection{Fine-tuning Results}
\begin{figure}
\centering
\includegraphics[width=0.5\linewidth]{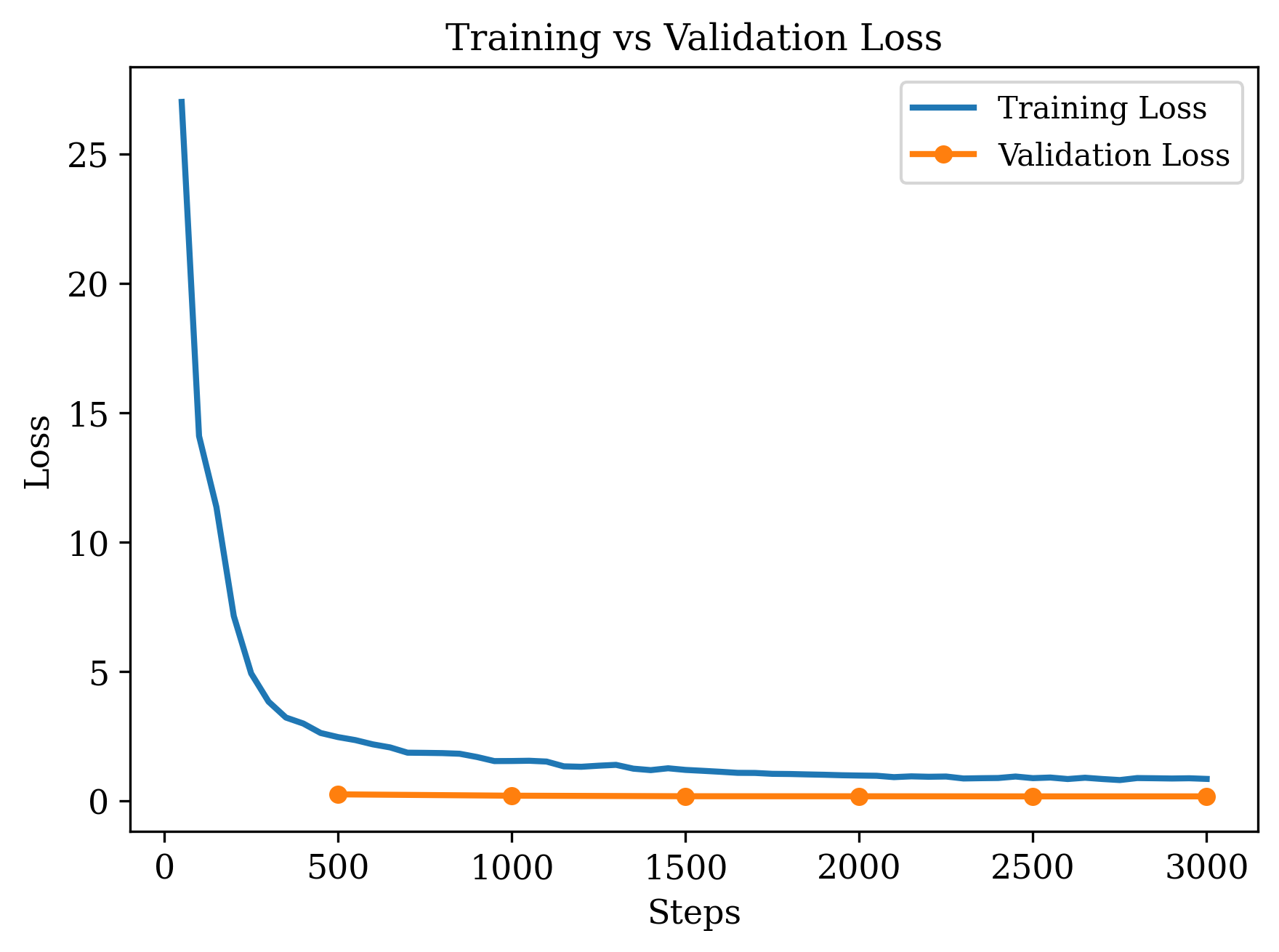}
\caption{Whisper-Small fine-tuning results for Assamese showing training and validation loss progression.}
\label{TrainingLoss}
\end{figure}

This section presents the quantitative results of the fine-tuning experiments. The fine-tuning logs presented in Table \ref{tab:training_log_full} demonstrate a robust numerical stability and efficient convergence over 13.64 epochs. Rapid improvement occurs in the initial steps (Steps 0 to 500) as training loss improves from 27.0038 to 2.4599 demonstrating that the model is adapting to phonetic of Assamese language. This assertion is additionally backed by the stability of the gradient norm, which, following an initial adaptation peak of 44.7003, stabilized within a consistent range of 4.0 to 8.0 for the duration of the session, signifying that neither vanishing nor exploding gradient issues were present. The use of a cosine learning rate scheduler was crucial. By reducing the learning rate from an initial value of 1.0×10−5 to a final value of 2.84×10−12, the optimizer enabled high-precision weight updates during the last 500 steps. Figure \ref{TrainingLoss} plots the results of Training with respect to Validation Loss. The plot shows rapid improvement in training loss over the initial 500–1,000 steps.  This suggests that the model is successfully identifying primary patterns available in the data. Both training and validation loss flatten out towards the end of training. The validation loss reached its lowest value at 2500 steps. This indicates that the model has reached its ideal checkpoint, and training beyond this may lead to overfitting.

\begin{table*}
\centering
\caption{Detailed Training Dynamics for Assamese Whisper-Small (50-Step Intervals)}
\label{tab:training_log_full}
\small
\setlength{\tabcolsep}{4pt}

\begin{tabular*}{\textwidth}{@{\extracolsep{\fill}}ccccc|ccccc@{}}
\toprule
\textbf{Step} & \textbf{Epoch} & \textbf{Loss} & \textbf{LR} & \textbf{Grad Norm} &
\textbf{Step} & \textbf{Epoch} & \textbf{Loss} & \textbf{LR} & \textbf{Grad Norm} \\
\midrule

50 & 0.23 & 27.0038 & 9.80e-06 & 44.7003 & 1550 & 7.05 & 1.1568 & 4.87e-06 & 5.7797 \\
100 & 0.45 & 14.0926 & 9.99e-06 & 14.5839 & 1600 & 7.27 & 1.1199 & 4.61e-06 & 6.0256 \\
150 & 0.68 & 11.3490 & 9.97e-06 & 18.1376 & 1650 & 7.50 & 1.0786 & 4.34e-06 & 6.0215 \\
200 & 0.91 & 7.1386 & 9.94e-06 & 12.6372 & 1700 & 7.73 & 1.0743 & 4.08e-06 & 3.9191 \\
250 & 1.14 & 4.9196 & 9.89e-06 & 27.2561 & 1750 & 7.95 & 1.0402 & 3.82e-06 & 5.1331 \\
300 & 1.36 & 3.8298 & 9.83e-06 & 9.7788 & 1800 & 8.18 & 1.0335 & 3.56e-06 & 5.9659 \\
350 & 1.59 & 3.2151 & 9.75e-06 & 8.5522 & 1850 & 8.41 & 1.0176 & 3.31e-06 & 4.8855 \\
400 & 1.82 & 2.9837 & 9.66e-06 & 10.3900 & 1900 & 8.64 & 1.0040 & 3.06e-06 & 6.7031 \\
450 & 2.05 & 2.6180 & 9.56e-06 & 8.5074 & 1950 & 8.86 & 0.9853 & 2.82e-06 & 4.8910 \\
500 & 2.27 & 2.4599 & 9.44e-06 & 8.6658 & 2000 & 9.09 & 0.9740 & 2.58e-06 & 5.0295 \\
550 & 2.50 & 2.3431 & 9.31e-06 & 7.5885 & 2050 & 9.32 & 0.9655 & 2.35e-06 & 5.1714 \\
600 & 2.73 & 2.1793 & 9.17e-06 & 10.1753 & 2100 & 9.55 & 0.9125 & 2.13e-06 & 6.9887 \\
650 & 2.95 & 2.0604 & 9.02e-06 & 7.8356 & 2150 & 9.77 & 0.9427 & 1.92e-06 & 6.6085 \\
700 & 3.18 & 1.8575 & 8.85e-06 & 8.0096 & 2200 & 10.00 & 0.9268 & 1.71e-06 & 5.0360 \\
750 & 3.41 & 1.8505 & 8.68e-06 & 5.3001 & 2250 & 10.23 & 0.9361 & 1.52e-06 & 4.1370 \\
800 & 3.64 & 1.8410 & 8.49e-06 & 5.9688 & 2300 & 10.45 & 0.8605 & 1.33e-06 & 4.7752 \\
850 & 3.86 & 1.8160 & 8.30e-06 & 6.9867 & 2350 & 10.68 & 0.8712 & 1.15e-06 & 4.4659 \\
900 & 4.09 & 1.6903 & 8.09e-06 & 5.5991 & 2400 & 10.91 & 0.8788 & 9.90e-07 & 5.4036 \\
950 & 4.32 & 1.5327 & 7.88e-06 & 6.4959 & 2450 & 11.14 & 0.9359 & 8.36e-07 & 5.6846 \\
1000 & 4.55 & 1.5357 & 7.66e-06 & 6.0020 & 2500 & 11.36 & 0.8731 & 6.95e-07 & 4.3694 \\
1050 & 4.77 & 1.5454 & 7.43e-06 & 4.7014 & 2550 & 11.59 & 0.8970 & 5.66e-07 & 5.6891 \\
1100 & 5.00 & 1.5144 & 7.19e-06 & 6.0267 & 2600 & 11.82 & 0.8393 & 4.49e-07 & 4.1134 \\
1150 & 5.23 & 1.3294 & 6.95e-06 & 6.1790 & 2650 & 12.05 & 0.8892 & 3.45e-07 & 6.2324 \\
1200 & 5.45 & 1.3127 & 6.70e-06 & 5.9978 & 2700 & 12.27 & 0.8398 & 2.55e-07 & 4.7803 \\
1250 & 5.68 & 1.3543 & 6.45e-06 & 7.0716 & 2750 & 12.50 & 0.7992 & 1.78e-07 & 4.5038 \\
1300 & 5.91 & 1.3869 & 6.19e-06 & 5.9283 & 2800 & 12.73 & 0.8765 & 1.14e-07 & 5.0079 \\
1350 & 6.14 & 1.2396 & 5.93e-06 & 4.8495 & 2850 & 12.95 & 0.8691 & 6.45e-08 & 3.7412 \\
1400 & 6.36 & 1.1833 & 5.67e-06 & 4.9130 & 2900 & 13.18 & 0.8594 & 2.89e-08 & 4.8757 \\
1450 & 6.59 & 1.2549 & 5.40e-06 & 5.0567 & 2950 & 13.41 & 0.8687 & 7.37e-09 & 5.2627 \\
1500 & 6.82 & 1.1917 & 5.14e-06 & 4.6896 & 3000 & 13.64 & 0.8440 & 2.84e-12 & 4.3517 \\

\bottomrule
\end{tabular*}
\end{table*}

\begin{table*}[t]
\centering
\caption{Similarity, accuracy, and structural metrics for fine-tuned Whisper-Small.}
\label{tab:whisper_accuracy_metrics}
\small
\setlength{\tabcolsep}{10pt}

\begin{tabular*}{\textwidth}{@{\extracolsep{\fill}}ccccccc@{}}
\toprule
\textbf{Step} & \textbf{Epoch} & \textbf{Validation Loss} & \textbf{WER} & \textbf{CER} & \textbf{MER} & \textbf{WIL} \\
\midrule
500  & 2.27 & 0.2473 & 0.6084 & 0.1785 & 0.5782 & 0.7555 \\
1000 & 4.55 & 0.1951 & 0.5129 & 0.1439 & 0.4916 & 0.6658 \\
1500 & 6.82 & 0.1744 & 0.4682 & 0.1322 & 0.4472 & 0.6185 \\
2000 & 9.09 & 0.1726 & 0.4533 & 0.1266 & 0.4383 & 0.6036 \\
2500 & 11.36 & 0.1689 & 0.4428 & 0.1246 & 0.4261 & 0.5930 \\
3000 & 13.64 & 0.1693 & 0.4407 & 0.1248 & 0.4237 & 0.5902 \\
\bottomrule
\end{tabular*}
\end{table*}

\begin{figure*}
\centering

\begin{subfigure}{0.48\textwidth}
    \centering
    \includegraphics[width=\linewidth]{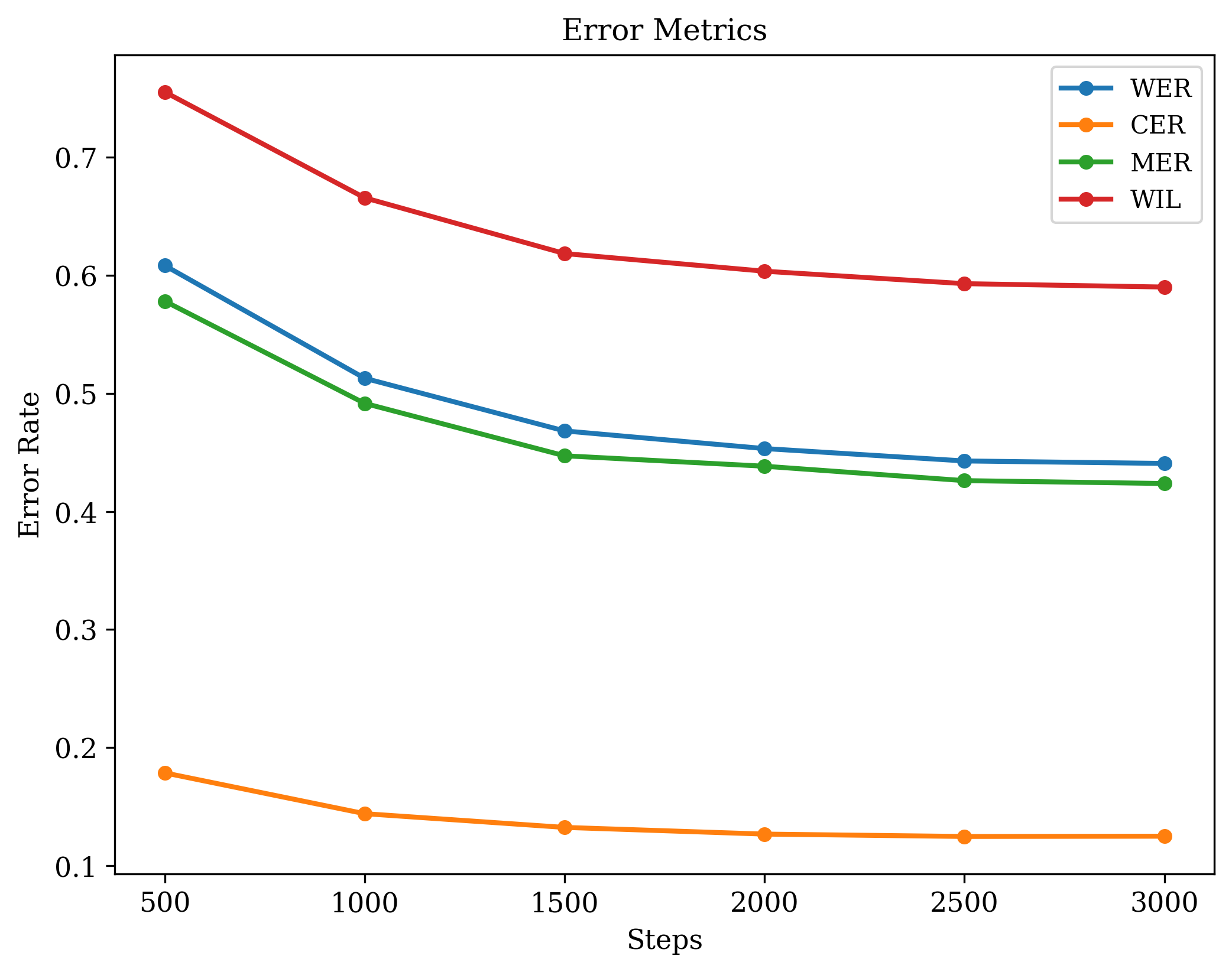}
    \caption{Error metrics}
    \label{fig:ft1a}
\end{subfigure}
\hfill
\begin{subfigure}{0.48\textwidth}
    \centering
    \includegraphics[width=\linewidth]{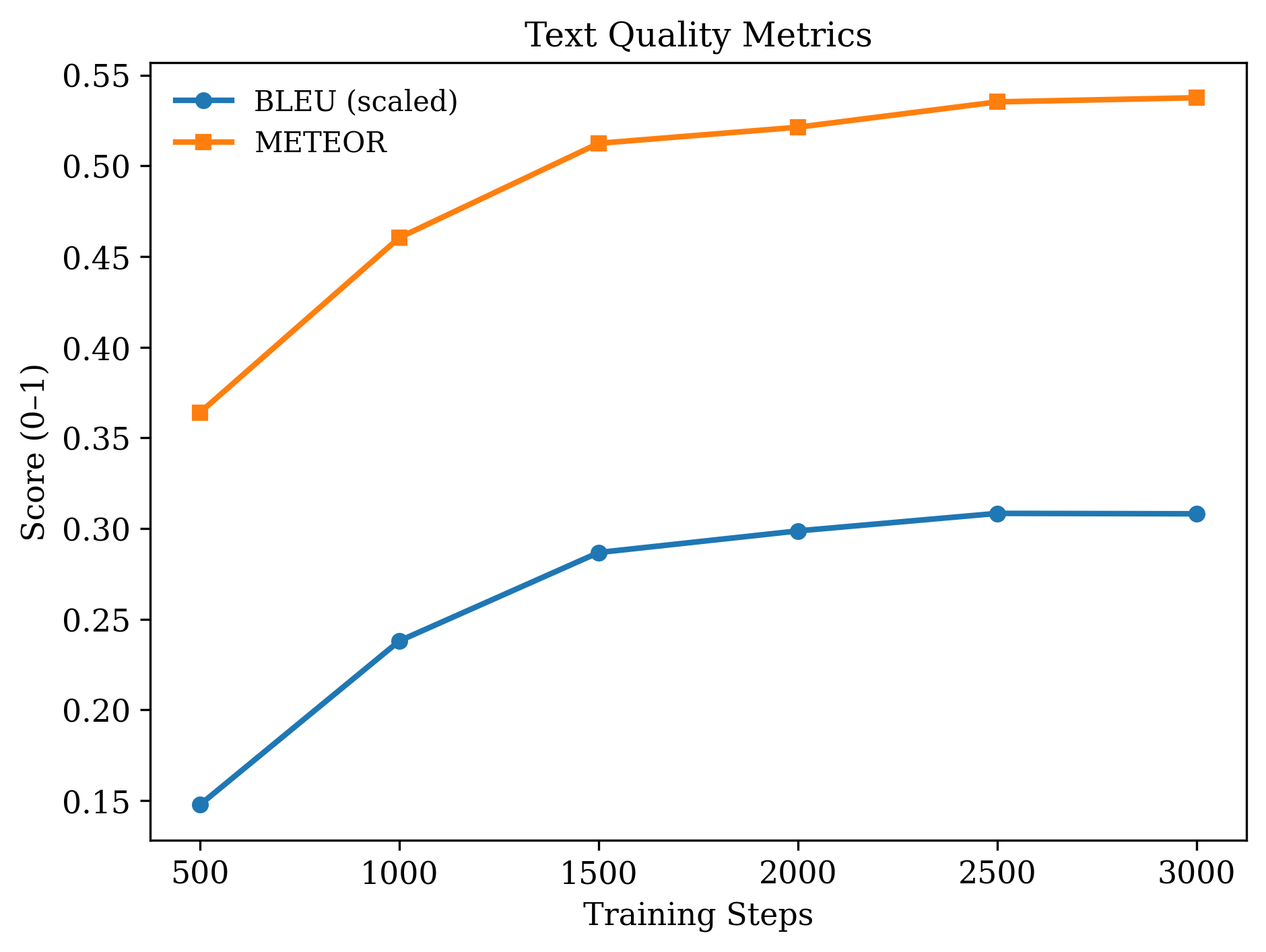}
    \caption{Text generation metrics}
    \label{fig:ft1b}
\end{subfigure}

\caption{Whisper-Small Fine-Tuning Results (Assamese)}
\label{fig:finetunningresults1}

\end{figure*}

\begin{figure*}
\centering

\begin{subfigure}{0.48\textwidth}
    \centering
    \includegraphics[width=\linewidth]{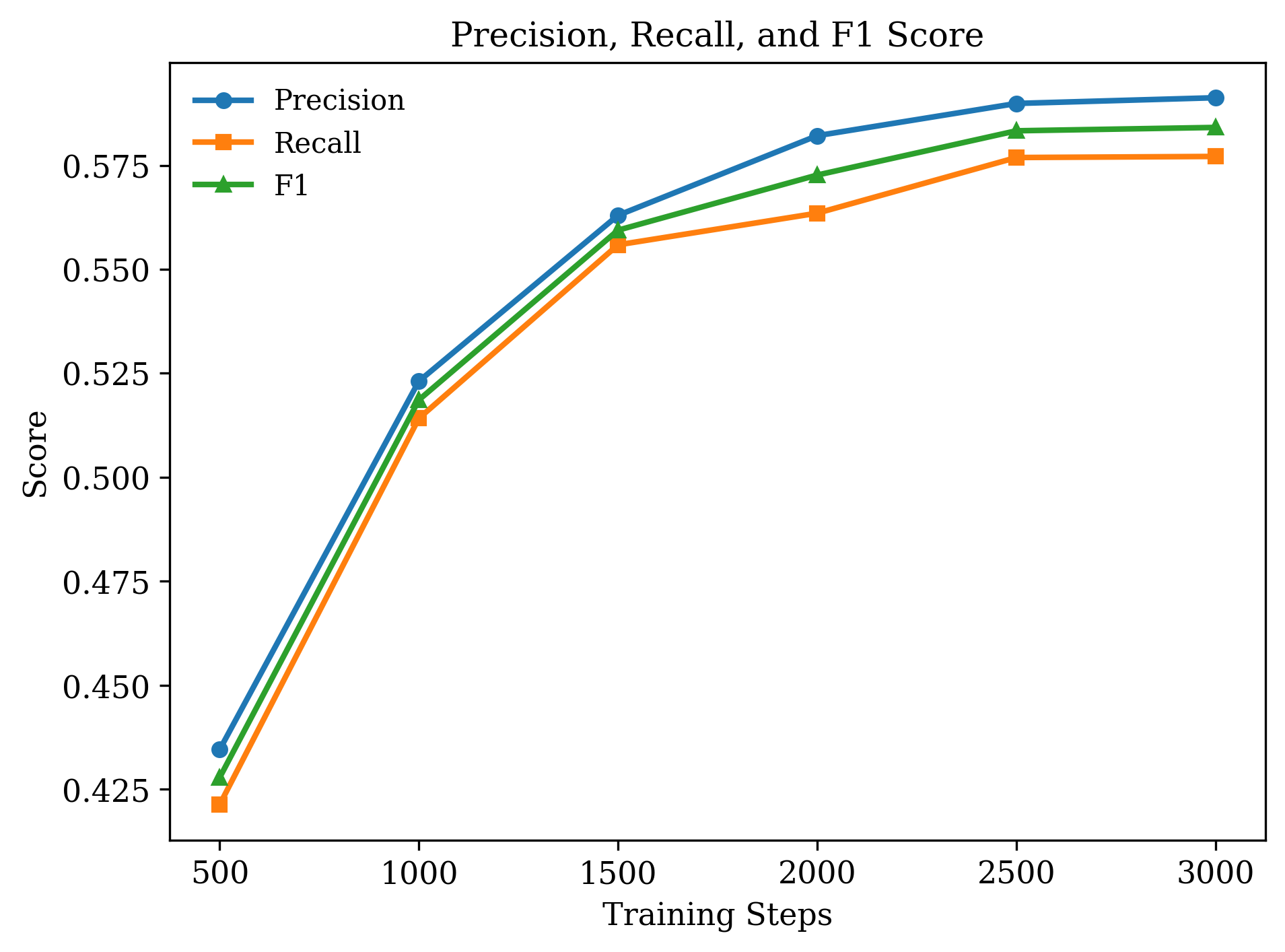}
    \caption{Precision, Recall, F1-score Metrics}
    \label{fig:ft2a}
\end{subfigure}
\hfill
\begin{subfigure}{0.48\textwidth}
    \centering
    \includegraphics[width=\linewidth]{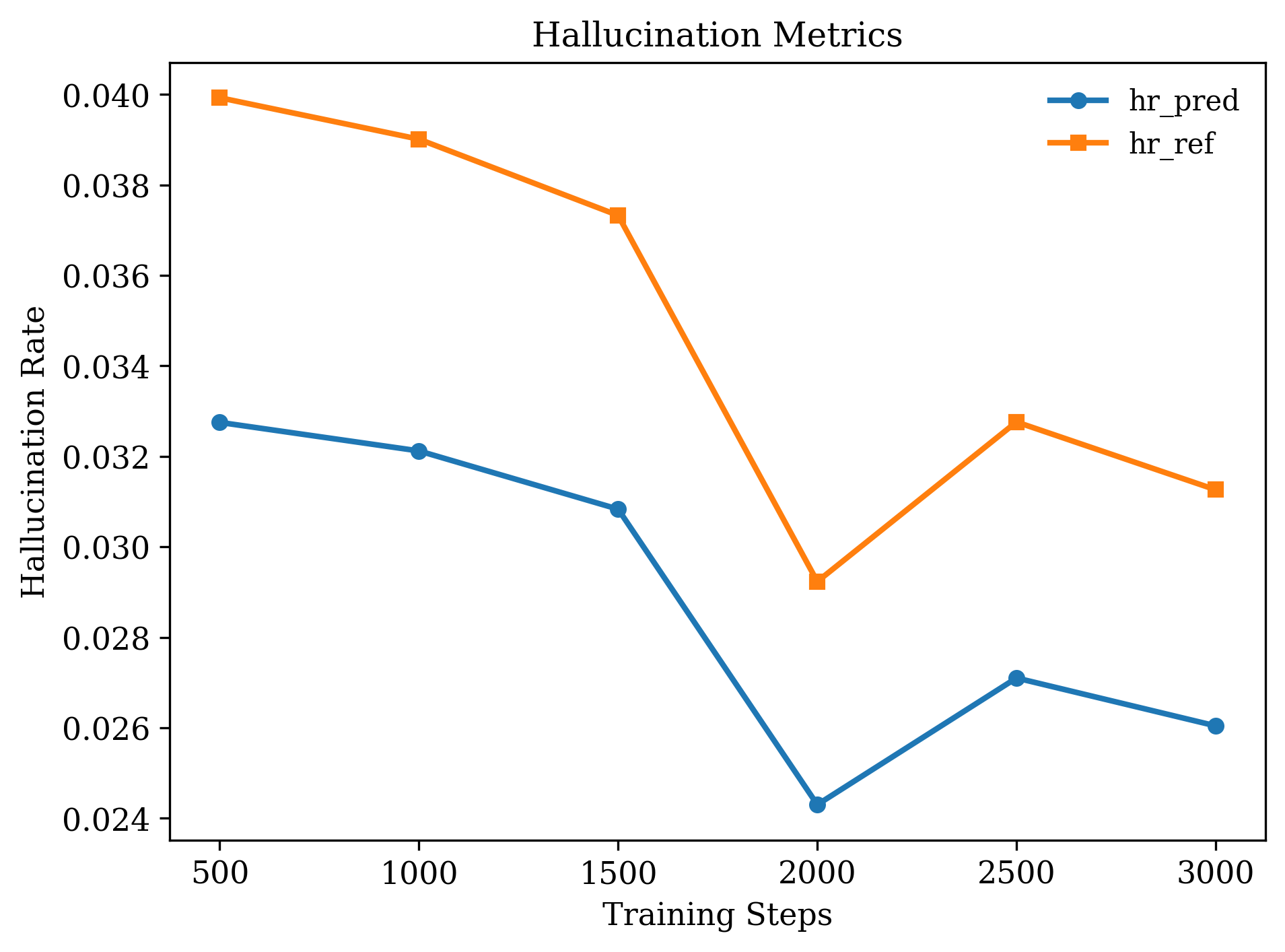}
    \caption{Hallucination Metrics}
    \label{fig:ft2b}
\end{subfigure}

\caption{Whisper-Small Fine-Tuning Results (Assamese)}
\label{fig:finetunningresults2}

\end{figure*}

The fine-tuning process was monitored using the training and validation loss every 500 steps, along with Word Error Rate (WER), Character Error Rate (CER), Match Error Rate (MER), BLEU Score, METEOR, Word Information Lost (WIL), Precision, Recall, F1-score, and Hallucination Error Rate (HER) on \texttt{dev.tsv}. The \texttt{Whisper-small} model was trained for 3000 steps, and the finetuning result is presented in Table~\ref{tab:whisper_accuracy_metrics} and Table \ref{tab:whisper_semantic_metrics}.
Improvements in transcription accuracy metrics, such as Word Error Rate (WER), Character Error Rate (CER), Match Error Rate (MER), and Word Information loss (WIL), shown in Figure \ref{fig:ft1a}, indicate that the model improves rapidly during the initial 2000 steps. A similar improvement is also observed in linguistic quality metrics, such as BLEU and METEOR, in Figure \ref{fig:ft1b}. Classification-based metrics during the training process are depicted in Figure \ref{fig:ft2a} while improvements in hallucination are shown in Figure \ref{fig:ft2b}. The final evaluation of the \texttt{test.tsv} samples from the Common Voice 24.0 Assamese subset yields a character error rate (CER) of 13.18\% and a word error rate (WER) of 43.75\%. Improvements in semantic and linguistic quality are supported by a BLEU score of 30.81 and a METEOR score of 0.5262. The F1 score of 0.5826 indicates a balanced performance in capturing relevant linguistic tokens.

\begin{table}
\centering
\caption{Semantic and Structural Performance Metrics for Assamese Whisper-Small Fine-Tuning}
\label{tab:whisper_semantic_metrics}
\begin{tabular}{@{}cccccc@{}}
\toprule
\textbf{Step} & \textbf{BLEU} & \textbf{F1} & \textbf{METEOR} & \textbf{HR$_{Pred}$} & \textbf{HR$_{Ref}$} \\ \midrule
500  & 14.78 & 0.4278 & 0.3640 & 0.0328 & 0.0399 \\
1000 & 23.80 & 0.5186 & 0.4604 & 0.0321 & 0.0390 \\
1500 & 28.68 & 0.5594 & 0.5124 & 0.0308 & 0.0373 \\
2000 & 29.87 & 0.5727 & 0.5213 & 0.0243 & 0.0292 \\
2500 & 30.84 & 0.5833 & 0.5352 & 0.0271 & 0.0328 \\
3000 & 30.81 & 0.5841 & 0.5375 & 0.0260 & 0.0313 \\ 
\bottomrule
\end{tabular}
\end{table}

\subsubsection{Comparison of the Fine-tuned Model with the Pre-trained Model}
The baseline performance of \texttt{Whisper-small} model in a Zero-shot setting yielded a Word Error Rate (WER) and a Character Error Rate (CER) of 2.0127 and 1.9091, respectively. The model produced incorrect transcriptions, rendering its performance unusable. An error rate exceeding 1.0 clearly indicates that the model is suffering from severe hallucination. Severity of hallucination is further confirmed by the HER$_{pred}$ of  0.5552. In the Zero-Shot state, scores for BLEU, METEOR, and F1 were effectively zero, indicating no overlap with the reference text. Match Error Rate (MER) and Word Information Lost (WIL) of 1.0000 indicate a complete failure in transcription, and no information from the reference was successfully captured by the model.

\begin{table*}
\caption{Comparison of Zero-shot with the Fine-tuned Model.}
\centering
\footnotesize
\setlength{\tabcolsep}{6pt}
\renewcommand{\arraystretch}{1.15}

\begin{tabular*}{\textwidth}{@{\extracolsep{\fill}}lcccc}
\toprule
\textbf{Metric} & \textbf{Zero-shot} & \textbf{Fine-tuned Whisper-Small} & \textbf{Absolute Reduction} & \textbf{Improvement (\%)} \\
\midrule
WER & 2.0127 & 0.4375 & 1.5752 & 78.26\% \\
CER & 1.9091 & 0.1318 & 1.7773 & 93.10\% \\
MER & 1.0000 & 0.4300 & 0.5700 & 57.00\% \\
WIL & 1.0000 & 0.6481 & 0.3519 & 35.19\% \\
BLEU & 0.0000 & 30.8100 & 30.8100 & $\infty$ \\
METEOR & 0.0000 & 0.5262 & 0.5262 & $\infty$ \\
F1 Score & 0.0000 & 0.5826 & 0.5826 & $\infty$ \\
Precision & 0.0000 & 0.5960 & 0.5960 & $\infty$ \\
Recall & 0.0000 & 0.5698 & 0.5698 & $\infty$ \\
HER$_{pred}$: & 0.5552 & 0.0183 & 0.5369 & 96.70\%\\
RTF & 0.2943 & 0.1990 & 0.0953 & 32.38\% \\
\bottomrule
\end{tabular*}

\label{zeroshotcompare}
\end{table*}

\begin{figure}
\centering
\includegraphics[width=0.5\columnwidth]{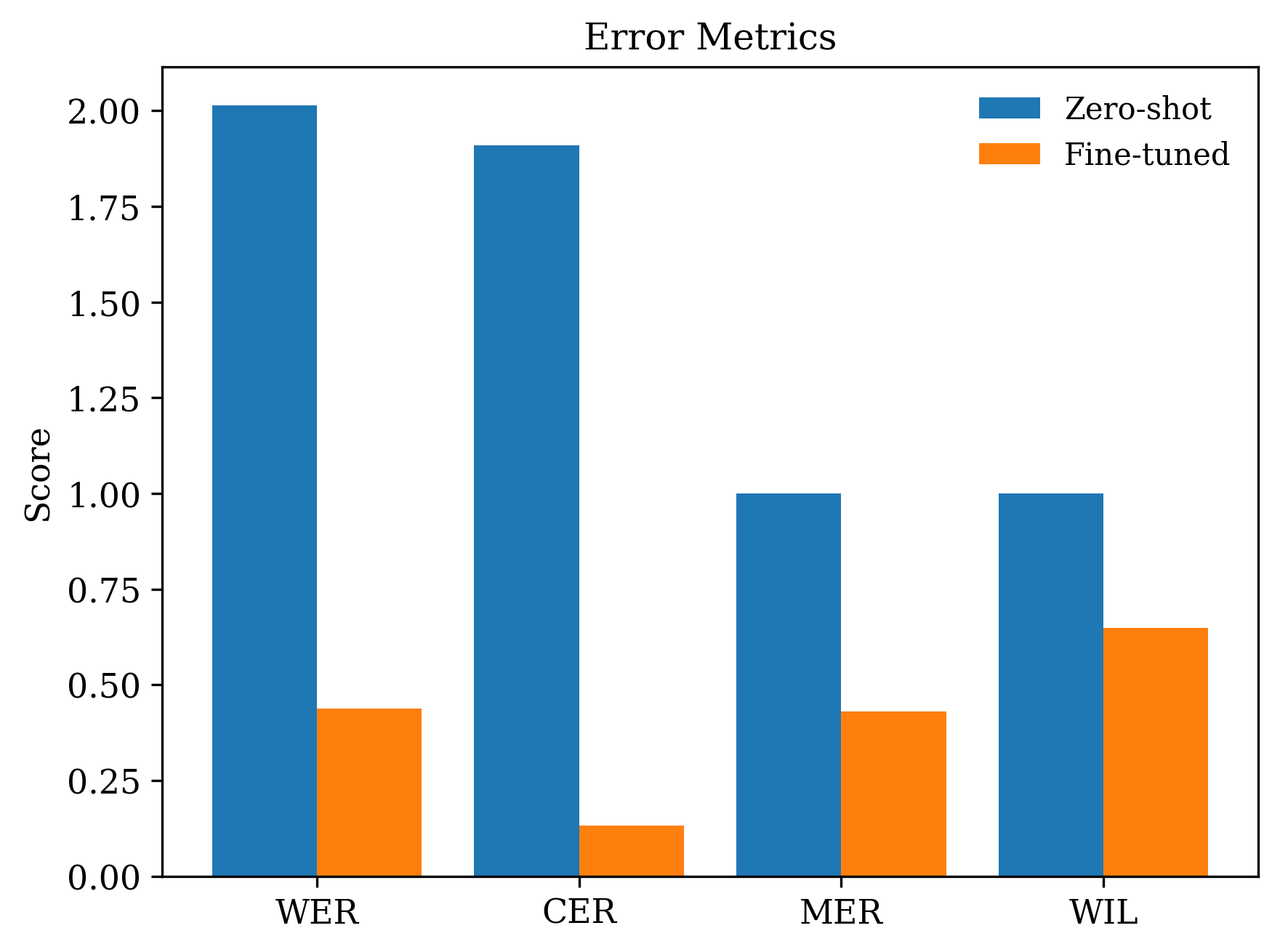}
\caption{Performance Comparison of Zero-Shot (baseline) and Fine-tuned Whisper-Small model.}
\label{comparison1}
\end{figure}

\begin{figure}
\centering
\includegraphics[width=0.5\columnwidth]{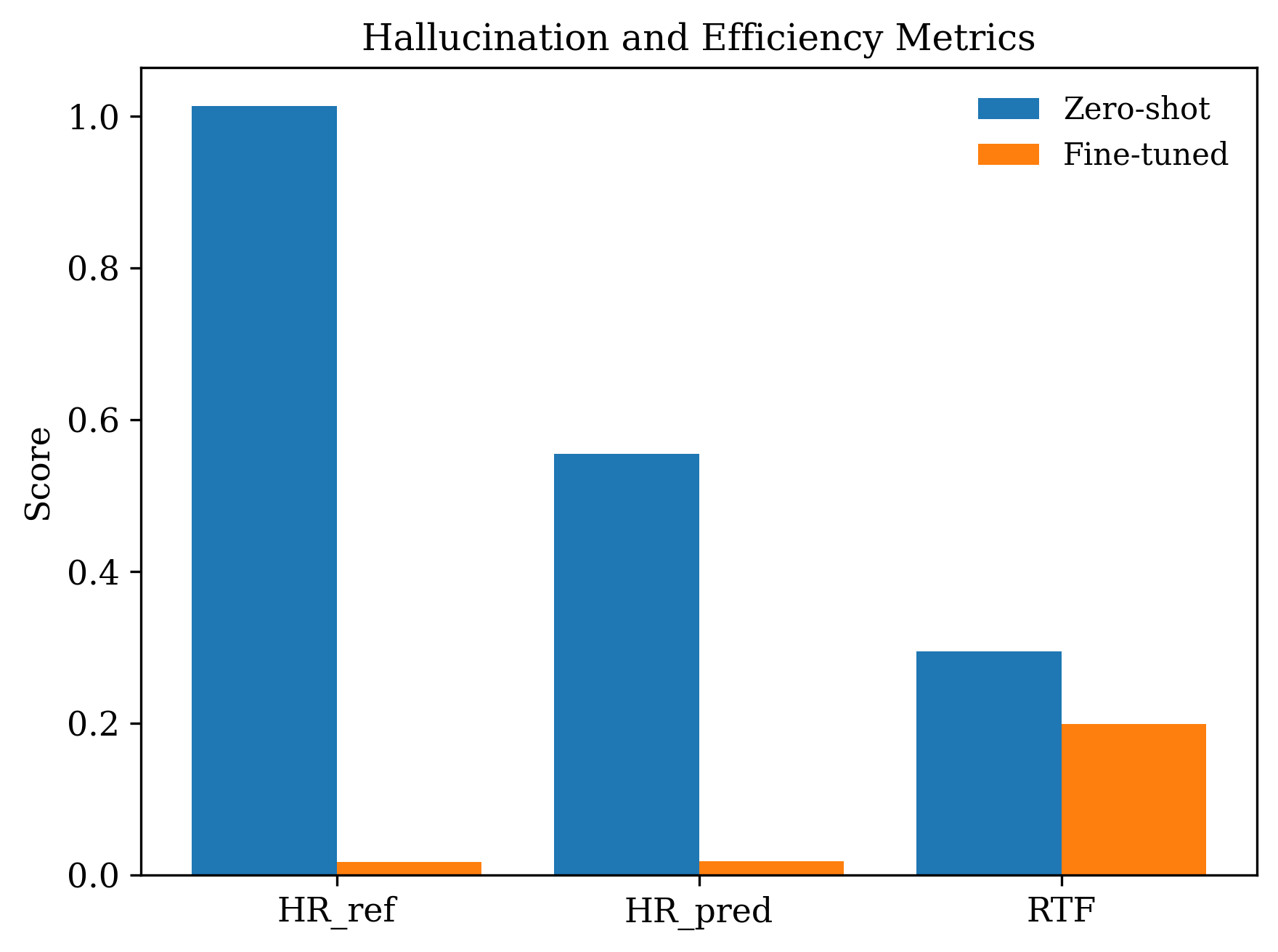}
\caption{Performance Comparison of Zero-Shot (baseline) and Fine-tuned Whisper-Small model.}
\label{comparison2}
\end{figure}

Table \ref{zeroshotcompare} shows the model's transformation after fine-tuning. The transition of CER from 1.9091 to 0.1318 represents a fundamental shift in the model's ability to map acoustic to phonetic representations in the Assamese language. This 93.10\% relative improvement in CER indicates that our approach to controlled fine-tuning is essential for high-fidelity ASR in Assamese. The fine-tuned model yielding a WER of 0.4375 may not be impressive, but it achieves a relative improvement of 78.26\% over the zero-shot baseline, demonstrating a significant improvement. In the Zero-Shot evaluation, the scores for BLEU, METEOR, and F1 were nearly zero, signifying no correspondence with the reference text. After the fine-tuning process, the model achieved a BLEU score of 30.8100 and an F1 score of 0.5826, demonstrating its improved capacity to produce semantically consistent sequences. The bar chart in Figure \ref{comparison1} clearly demonstrates this transition from a non-functional Zero-Shot (baseline) to a specialized Assamese ASR in terms of WER, CER, MER and WIL. A Significant reduction of 96.70\% in hallucination error rate, as well as 32.38\% improvement in efficiency of transcription, measured using real-time factor, is depicted in Figure \ref{comparison2}.

\subsubsection{Evaluation Metric Analysis}
Fine-tuned Whisper-small produces a significant improvement over the zero-shot evaluation with WER and CER of 43.75\% and 13.18\%, respectively. However, a substantial 30.51\% gap remains between WER and CER. This gap between CER and WER is a classic characteristic of low-resource agglutinative languages \citep{bekarystankyzy2025integrated}. A lower CER of 13.18\% is a phonetic success, indicating that the model can accurately map the Assamese scripts to their corresponding acoustic representations. But the model is still unable to understand the complex grammar of Assamese language. From the generated samples, it appears the model correctly identifies the root words but struggles with the suffix. For example, the root word ``{\assamese{পাহাৰ}}'' (\textipa{[pa.ha\textturnr]}) is correctly identified, while it struggles with sufix `{\assamese{ত}}'(\textipa{[\textsubbridge{t}]}) in ``{\assamese{পাহাৰত}}'' (\textipa{[pa.ha.\textturnr O\textsubbridge{t}]}).
Additionally, in several instances, the model correctly identifies root words but struggles with word boundaries, where one word ends and a new one begins (\textit{see} sample 140, Table \ref{MorphologicalErros}). Additionally, some words with similar acoustic sounds are produced despite their different visual representations, leading to non-existent words in  Assamese (\textit{see} sample 57, Table \ref{OrthographicErrosGS}). Another reason for the high Word Error Rate (WER) is its mathematical sensitivity. WER evaluates errors at the word level, meaning that if even a single character in a word is predicted incorrectly, the entire word is counted as wrong. As a result, minor spelling mistakes can significantly increase the WER value. In contrast, Character Error Rate (CER) measures errors at the character level. Therefore, when a character is predicted incorrectly, only that specific character is marked as an error rather than the whole word. Because of this difference, WER is generally more sensitive to small prediction errors, while CER provides a more fine-grained evaluation of transcription accuracy in any speech recognition model, and it is clearly justified in above analysis of the Whisper-small model performance. The fine-tuned model yields MER and WIL of 0.4300 and 0.6481, respectively. The difference between MER and WIL stems from the fact that MER measures the overall mismatch proportion, whereas WIL assesses the statistical independence between the two strings. Linguistic and semantic similarity are assessed using the BLEU and METEOR evaluation metrics. The proposed fine-tuned approach achieved a BLEU score of 30.81 and a METEOR score of 0.5262. BLEU focuses on precision, measuring how many of the predicted phrases are present in the reference. In contrast, METEOR emphasizes alignment and imposes penalties on matches that are fragmented. 

\subsection{Qualitative Analysis} \label{error}
The quantitative analysis in Section 7.1 provided a high-level view of our fine-tuned model's performance. To understand the model's performance at the granular level, two types of qualitative analysis, viz., \textit{Orthographic error} and \textit{Morphological error}, are examined on the predicted transcripts given in Table  \ref{qualitative}. 
These samples correspond to the same reference script used for the zero-shot evaluation of the model presented in Table \ref{samplepredzero}

\textit{Orthographic} and \textit{morphological} errors in Assamese arises due to a combination of phonetic similarity, orthographic ambiguity, retroflex–dental and aspirated–un-aspirated confusions, sibilant mergers, and grapheme–phoneme mismatches. Orthographic errors refer to discrepancies in graphetic and phonetic shifts, where the predicted word is phonetically correct or closely resembles the intended pronunciation, but its orthographic representation deviates from the standard spelling. In such cases, the model generates a word that sounds similar to the reference word but differs visually or textually. Confusion in the use of vowels (matras), Consonant aspiration or voicing, and homophonic substitutions are common orthographic errors in Assamese. Morphological Errors occur when the model correctly identifies the root word, but is unable to construct the grammatical structure. This typically involves the omission, addition, or substitution of suffixes, prefixes case markers, or other inflectional morphemes, including incorrect word boundary segmentation. 
A detailed analysis of the model for \textit{Orthographic} and \textit{Morphological} errors is discussed in the following sections

\begin{table*}[t]
\caption{Qualitative comparison of ground-truth reference (REF) and model predictions (PRED)}
\label{qualitative}
\centering
\small
\renewcommand{\arraystretch}{1.2}

\begin{tabular}{p{1.2cm} p{0.85\textwidth}}
\toprule
\textbf{Sample No.} &
\textbf{Ground-truth reference (REF) and Model predictions (PRED) Outcomes} \\
\midrule

5 & 
REF:{\assamese{কোৱা হয় সেই পীঠৰ মন্দিৰ সৰ্ব্ব প্ৰথম নিৰ্মাণ কৰিছিল নৰকাসুৰে}}\\ 
& PRED:{\assamese{কোৱা হয় সেই পিঠৰ মণ্ডিৰ শ্বৰ্বপ্ৰথম নিৰ্মাণ কৰিছিল নৰক আশুৰে}}\\ 
\midrule
12 & 
REF: {\assamese{কন্দা দেখি মাকৰ কোলাত হেৰি উঠিলগৈ}}\\
& PRED: {\assamese{খন্দা দেখি মাকৰ কলাত হেৰি উঠিলকৈ}}\\
\midrule

41 & 
REF:{\assamese{লৰাজনক সোধাত গম পালে তাৰ নাম পবিত্ৰ নাথ}}\\
& PRED:{\assamese{লৰাজনক সোধাত গম পালে তাৰ নাম পবিত্ৰ নাথ}}\\
\midrule

52 & 
REF:{\assamese{এবছৰ আগেয়ে কানাডালৈ পলাই গৈছিল}}\\
& PRED:{\assamese{এবছৰ আগেয়ে কাণাডালৈম পলায়ে গৈছিল }}\\
\midrule
57& 
REF: {\assamese{ ঘৰৰ গৃহিণীৰ ওপৰতে গোটেই পৰিয়ালৰ সুখ দুখ নিৰ্ভৰ কৰে}}\\
& PRED: {\assamese{ঘৰৰ গৃঘিনীৰ ওপৰতে গোটেই পৰিয়ালৰ শুখটোক নিৰ্পাৰ কৰে}}\\
\midrule
71 & 
REF:{\assamese{মণ্ডলৰ ঘৰৰ পৰা খবৰ আহিল দৰা এতিয়াও ভাল হোৱা নাই}}\\
& PRED:{\assamese{মণ্ডলৰ ঘৰৰ পৰা খবৰ আহিল ধৰাই তিয়াও ভাল হোৱা নাই}}\\
\midrule
75 & 
REF:{\assamese{সি চাগে তোমাৰ কথা শুনা নাই আগতে	}}\\
& PRED:{\assamese{সি চাগে তোমাৰ কথা শুনা নাই আগতে}}\\
\midrule
114 & 
REF:{\assamese{আগতে বন্দবস্ত হৈছিল আপোনালোকৰ ফালৰ মদন মণ্ডলৰ ভায়েকৰ লগত}}\\
& PRED:{\assamese{আগতে বন্দৱস্ত হৈছিল আপ্নালোকৰ ফালৰ মদন মণ্ডলৰ ভায়কৰ লগত}}\\
\midrule
140 & 
REF:{\assamese{কাণ্ডকাৰখানা দেখি লভিতা আৰু লগৰ ছোৱালী কেজনী নিজৰ ঘৰৰ দুৱাৰমুখত থুপ খালেগৈ}}\\
& PRED:{\assamese{কাণ্ডকাৰখনা দেখি লভিতা আৰু লগৰ ছোৱালীকেইজনী মিজৰ ঘৰৰ দুৱাৰমুখত থোক খালেগৈ}}\\
\midrule
182 & 
REF: {\assamese{লভিতাহঁতে উত্তেজিত হৈ বাহিৰৰ ফালে চাবলৈ ধৰে}} \\
& PRED: {\assamese{লভিতাহঁতে উত্তেজিত হৈ বাহিৰৰ ফালে চাবলৈ ধৰে}}\\
\midrule
276 & 
REF:{\assamese{বিশেষকৈ শ্ৰীমান অভয় দুৱৰা আৰু শ্ৰীমান তৰুণ দুৱৰাৰ বাবেইহে}}\\
& PRED:{\assamese{বিশেষকৈ শ্ৰীমান অভয় দুৱৰা আৰু শ্ৰীমান তৰুণ দুৱৰাৰ বাবেইহে}}\\
\midrule
327&
REF: {\assamese{কোৰআন শ্বৰীফ ইছলাম ধৰ্মৰ মূল ধৰ্মগ্ৰন্থ}}\\ 
& PRED: {\assamese{কুৰাণচৰিপ ইছলাম ধৰ্মৰ মোল ধৰ্মগন্থ}}\\
\midrule
345	&
REF: {\assamese{মানুহটোৱে কলে ভালকৈ শিক নিশিকিবি}}\\
& PRED: {\assamese{মানুহটোৱে কলে ভালকৈ শিক নিশিকিবি}}\\
\midrule
377 & 
REF:{\assamese{তহঁতৰ নো খায় কোনে}}\\
& PRED:{\assamese{তহঁতৰ দু খাই কোনে}}\\
\midrule
379 & 
REF:{\assamese{মাষ্টাৰ গলেই হয়}}\\
& PRED:{\assamese{মাষ্টৰ গলেই হয়}}\\
\bottomrule
\end{tabular}

\end{table*}

\subsubsection{Orthographic Error Analysis}
Analysis of the evaluated samples revealed that the predicted transcripts exhibit frequent vowel substitution errors, particularly among phonetically similar Assamese vowel diacritics. The most prominent confusions are observed between `{\assamese{ু}}'([u]), `{\assamese{ূ}}' ([u:]), and `{\assamese{ো}}' (\textipa{[U]}), as well as between `{\assamese{ি}}' ([i]) and `{\assamese{ী}}' ([i:]), leading to vowel substitution errors. For instance, in Sample 327 in Table \ref{OrthographicErrosVS}, the vowel `{\assamese{ো}}' associated with the character `{\assamese{ক}}' is misrecognized as `{\assamese{ু}}', resulting in the predicted word ``{\assamese{কুৰাণ}}'' (\textipa{[ku.\textturnr an]}) instead of the reference word ``{\assamese{কোৰআন}}'' (\textipa{[kU.\textturnr a.an]}). Another vowel substitution error  is seen in Sample 5, where the vowel `{\assamese{ী}}' associated with the character `{\assamese{প}}' ([p]) is incorrectly recognized as `{\assamese{ি}}` resulting the predicted word as ``{\assamese{পিঠৰ}}'' (\textipa{[pi.t\textsuperscript{h}O\textturnr]}) instead of ``{\assamese{পীঠৰ}}'' (\textipa{[pi:.t\textsuperscript{h}O\textturnr]} ). Such kind of substitution errors among vowel pairs viz.,\{(`{\assamese{ো}}', `{\assamese{ু}}'), (`{\assamese{ী}}', `{\assamese{ি}}'), (`{\assamese{ূ}}',`{\assamese{ো}}')\} are observed in samples 327, 140, and 377. Vowel omission errors are also seen in some instances as shown in Table \ref{OrthographicErrosVS}. For example, in Sample 12 and 114, the `{\assamese{ ো }}'is omitted for the words ``{\assamese{ কোলাত}}'' (\textipa{[kU.lat]}) and ``{\assamese{আপোনালোকৰ}}'' (\textipa{[a.pU.na.lU.kO\textturnr]}). Similarly, `{\assamese{ া }}' ([a])  were missing in Sample 379 and 57, whereas `{\assamese{ ে }}' ([e]) is omitted in Samples 114. These patterns indicate that the model struggles to accurately capture temporal acoustic cues, leading to errors in grapheme representation. 


\begin{table*}
\caption{Orthographic Error Analysis (Vowel Substitution/Omission)}
\centering
\footnotesize
\setlength{\tabcolsep}{6pt}
\renewcommand{\arraystretch}{1.15}

\begin{tabular*}{\textwidth}{@{\extracolsep{\fill}}cllll}
\toprule
\textbf{Sample} & \textbf{REF Word} & \textbf{PRED Word} & \textbf{Vowel Substitution(s)} & \textbf{Vowel Omission(s)}\\
\midrule
5 & {\assamese{পীঠৰ}} & {\assamese{পিঠৰ}} & {\assamese{ ী}} $\rightarrow$ {\assamese{ি}}& --\\
12 & {\assamese{ কোলাত}} & {\assamese{ কলা}} & -- & {\assamese{ ো}}\\ 
114 & {\assamese{ ভায়েকৰ }} & {\assamese{ভায়কৰ}}&-- & {\assamese{  ে }} \\ 
114 & {\assamese{ আপোনালোকৰ}} & {\assamese{আপ্নলোকাৰ}} & -- &{\assamese{ো }} \\ 
140 & {\assamese{ থুপ}} & {\assamese{ থোক}} & {\assamese{ ু}} $\rightarrow$ {\assamese{ো}}& --\\
140 & {\assamese{কাণ্ডকাৰখানা}} & {\assamese{কাণ্ডকাৰখনা}} & --&{\assamese{া}}  \\ 
327 & {\assamese{কোৰআন}} & {\assamese{ কুৰাণ}} & {\assamese{ ো}} $\rightarrow$ {\assamese{ ু}} & --\\ 
327 & {\assamese{ শ্বৰীফ}} & {\assamese{ চৰিপ}} &   {\assamese{ী}} $\rightarrow$ {\assamese{ ি} }& --   \\ 
327 & {\assamese{ মূল}} & {\assamese{ মোল}} & {\assamese{ ূ}} $\rightarrow$ {\assamese{ ো}} & --\\ 
377 & {\assamese{ নো}} & {\assamese{দু}} &  {\assamese{  ো}}$\rightarrow$ {\assamese{  ু}}& -- \\ 
379 & {\assamese{ মাষ্টাৰ}} & {\assamese{ মাষ্টৰ}} &--& {\assamese{ া}} \\
\bottomrule
\end{tabular*}
\label{OrthographicErrosVS}
\end{table*}

Phonetically motivated grapheme (character) substitution errors are also observed in several samples, as shown in Table \ref{OrthographicErrosGS}. These errors refer to the replacement of a grapheme   by another grapheme based on phonetic similarity, and are not restricted to consonantal substitutions only. For instance, in Sample 327, the conjunct consonant grapheme `{\assamese{শ্ব}}' ([xw]) is replaced by `{\assamese{চ}}' (\textipa{[s]}) resulting the predicted word ``{\assamese{চৰিপ}}'' (\textipa{[s.\textturnr ip]}) instead of ``{\assamese{শ্বৰীফ}}'' ( \textipa{[xw.\textturnr if])}. In another instance, as shown in Sample 377, semi-vowel `{\assamese{য়}}' ([j]) is replaced by vowel `{\assamese{ই}}' ([i]), resulting in the prediction ``{\assamese{খাই}}'' (\textipa{[k\textsuperscript{h}ai]}) instead of the correct form ``{\assamese{খায়}}'' (\textipa{[k\textsuperscript{h}aj]}). 

\begin{table*}
\caption{Orthographic Error Analysis (Grapheme Substitution Errors}
\centering
\footnotesize
\setlength{\tabcolsep}{6pt}
\renewcommand{\arraystretch}{1.15}

\begin{tabular*}{\textwidth}{@{\extracolsep{\fill}}clll}
\toprule
\textbf{Sample} & \textbf{REF Word} & \textbf{PRED Word} & \textbf{Grapheme Substitution Found} \\
\midrule
5 & {\assamese{মন্দিৰ}} & {\assamese{মণ্ডিৰ}} & {\assamese{ন}} $\rightarrow$ {\assamese{ণ}}, {\assamese{দ}} $\rightarrow$ {\assamese{ড}}  \\
5 & {\assamese{সৰ্ব্ব}} & {\assamese{শ্বৰ্ব}} & {\assamese{ স}} $\rightarrow$ {\assamese{শ}} \\
5 & {\assamese{ নৰকাসুৰে}} & {\assamese{নৰক আশুৰে}} & {\assamese{ স}} $\rightarrow$ {\assamese{শ}} \\
12 & {\assamese{ কন্দা}} & {\assamese{ ঘণ্ডা}} & {\assamese{ ক}} $\rightarrow$ {\assamese{ ঘ, ন্দ}} $\rightarrow$ {\assamese{ ণ্ড}} \\
12 & {\assamese{ উঠিলগৈ}} & {\assamese{ উঠিলকৈ}} & {\assamese{গ}} $\rightarrow$ {\assamese{ক}} \\
52 & {\assamese{ কানাডালৈ}} & {\assamese{ কাণাডালৈ}} & {\assamese{ ন}} $\rightarrow$ {\assamese{ ণ}}\\
57 & {\assamese{ নিৰ্ভৰ}} & {\assamese{নিৰ্পাৰ}} & {\assamese{ ভ}} $\rightarrow$ {\assamese{ প}} \\
57 & {\assamese{ গৃহিণীৰ}} & {\assamese{ গৃহিনীৰ}} & {\assamese{ ণী}} $\rightarrow$ {\assamese{ নী}} \\
57 & {\assamese{ সুখ}} & {\assamese{ শুখ}} & {\assamese{ স}} $\rightarrow$ {\assamese{ শ}} \\
114 & {\assamese{ বন্দবস্ত}} & {\assamese{ বন্দৱস্ত}} & {\assamese{ ব}} $\rightarrow$ {\assamese{ ৱ}} \\
140 & {\assamese{ নিজৰ}} & {\assamese{ মিজৰ}} & {\assamese{ ন}} $\rightarrow$ {\assamese{ ম}} \\
140 & {\assamese{ থুপ}} & {\assamese{ থোক}} & {\assamese{ প}} $\rightarrow$ {\assamese{ ক}} \\
327 & {\assamese{কোৰআন}} & {\assamese{ কুৰাণ}} & {\assamese{ ন}} $\rightarrow$ {\assamese{ ণ}}\\
327 & {\assamese{ শ্বৰীফ}} & {\assamese{ চৰিপ}} & {\assamese{ শ্ব}} $\rightarrow$ {\assamese{ চ }},{\assamese{ফ }} $\rightarrow$ {\assamese{ প}}\\  
327& {\assamese{ ধৰ্মগ্ৰন্থ}} & {\assamese{ ধৰ্মগন্থ}} & {\assamese{গ্ৰ}} $\rightarrow$ {\assamese{গ}}\\
377 & {\assamese{ নো}} & {\assamese{দু}} & {\assamese{ ন }} $\rightarrow$ {\assamese{ দ}} \\
377 & {\assamese{ খায় }} & {\assamese{খাই }} & {\assamese{ য় }} $\rightarrow$ {\assamese{ ই}} \\
\bottomrule
\end{tabular*}
\label{OrthographicErrosGS}
\end{table*}

In Sample 57,  ``{\assamese{গৃহিণীৰ}}'' (\textipa{[g\textturnr i.Hi.ni:\textturnr]}) predicted as ``{\assamese{গৃঘিনীৰ}}''( \textipa{[g\textturnr i.g\textsuperscript{h}i.n:i\textturnr]}), and in Sample 125, ``{\assamese{পুৰণি}}'' (\textipa{[pu.\textturnr O.Ni]}) was rendered as ``{\assamese{পুৰণী}}'' (\textipa{[pu.\textturnr O.ni:]}), indicating that the model failed to distinguish between dental `{\assamese{ন}}' ([n]) and retroflex `{\assamese{ণ}}' (\textipa{[N]}). Although these substitutions are phonetically similar, they are orthographically incorrect. Assamese language uses several conjuncts, in which nearly similar sounds can be generated by different combinations of characters. ``{\assamese{মন্দিৰ}}'' (\textipa{[mOn.di\textturnr]}) and ``{\assamese{মণ্ডিৰ}}'' (\textipa{[mO\textsubdot{N}\textsubdot{d}i\textturnr]}) makes similar acoustic sound in Sample 5. While the former is a correct word from the reference text, the latter is not an acceptable spelling. Sibilant shifts are also observed in Sample 57 where ``{\assamese{সুখ}}'' (\textipa{[xuk\textsuperscript{h}]}) became ``{\assamese{শুখ}}'' (\textipa{[xuk\textsuperscript{h}]}), replacing `{\assamese{স}}' (\textipa{[x]}) with phonetically similar consonant `{\assamese{শ}}' (\textipa{[x]}). The same pattern is also observed in Sample 5. Labial inconsistencies are also observed in some samples. This type of error arises when labial sounds are pronounced incorrectly or substituted with other labial sounds. For instance, in Sample 57, ``{\assamese{নিৰ্ভৰ}}'' (\textipa{[ni\textturnr b\textsuperscript{H}O\textturnr]}) is transcribed as ``{\assamese{নিৰ্পাৰ}}'' (\textipa{[ni\textturnr pa\textturnr]}), by substituting `{\assamese{ভ}}' (\textipa{[b\textsuperscript{H}]}) with `{\assamese{প}}' (\textipa{[p]}). A similar pattern of error is also noticed in Sample 327, where `{\assamese{ফ}}' (\textipa{[f]}) is replaced by `{\assamese{প}}' (\textipa{[p]}). Aspiration confusion like `{\assamese{ক}}' (\textipa{[k]}) to `{\assamese{ঘ}}'(\textipa{[g\textsuperscript{H}]}),`{\assamese{দ}}' (\textipa{[d]}) to `{\assamese{ধ}}'(\textipa{[d\textsuperscript{H}]}), `{\assamese{দ}}' (\textipa{[d]}) to `{\assamese{ড}}' (\textipa{[\textsubdot{d}]}), {\assamese{ন}}' (\textipa{[n]}) to `{\assamese{দ}}' (\textipa{[d]}), and `{\assamese{প}}' (\textipa{[p]}) to {\assamese{ক}}' (\textipa{[k]}) are seen in Samples 12, 71, 5, 377, 140 respectively. Nasal inconsistencies are also observed in Samples 327, 57, 140, 327, 52, and 5. This type of error occurs due to the substitution between one nasal sound with another, such as {\assamese{ন}}' (\textipa{[n]}), {\assamese{ম}}' (\textipa{[m]}), and {\assamese{ণ}}' (\textipa{[N]}). Conjunct loss error is also found in Sample 327, where `{\assamese{গ্ৰ}}' (\textipa{[g\textturnr]}) formed by the conjuction of `{\assamese{গ}} (\textipa{[g]}) and {\assamese{ৰ}}' (\textipa{[\textturnr]}) is missing in the word ``{\assamese{ ধৰ্মগ্ৰন্থ}}'' (\textipa{[d\textsuperscript{H}O\textturnr .mO.g\textturnr On.t\textsuperscript{H}O]})  producing the transcribed word ``{\assamese{ ধৰ্মগন্থ}}'' (\textipa{[d\textsuperscript{H}O\textturnr .mO.gOn.t\textsuperscript{H}O]}). 
The above-discussed orthographic errors demonstrate that while the acoustic model achieved a high degree of phonetic literacy, it frequently misrepresents words visually due to limitations in its lexical knowledge. Detailed orthographic error analysis of all samples is reported in Table \ref{OrthographicErrosVS} and Table \ref{OrthographicErrosGS}.

\subsubsection{Morphological Error Analysis}

The qualitative analysis of the fine-tuned model's predicted outputs, presented in Table \ref{qualitative}, identified several morphological errors, viz., word boundary errors, affix-related errors, and lexical substitutions, and different categories of morphological errors found are presented in Table \ref{MorphologicalErros}.
The most commonly observed morphological errors involve correctly identifying word boundaries, particularly word merging and splitting.
For instance, in Sample 5, the correct phrase ``{\assamese{সৰ্ব্ব প্ৰথম}}'' (\textipa{[\text{xO\textturnr .bbO.p\textturnr O.t\textsuperscript{H}Om}]}) is incorrectly merged to form a single word, ``{\assamese{শ্বৰ্বপ্ৰথম}}'' (\textipa{[\text{xO\textturnr .bbO.p\textturnr O.t\textsuperscript{H}Om}]}). The same error is also observed for the words ``{\assamese{ছোৱালী কেজনী}}'' (\textipa{[sU.wa.li ke.zO.ni]}) and ``{\assamese{কোৰআন শ্বৰীফ}}'' (\textipa{[kU.\textturnr a.an xw.\textturnr if]}) in Samples 140 and 327, respectively. In contrast, word splitting error is observed in Sample 5, where the word ``{\assamese{নৰকাসুৰে}}'' (\textipa{[nO\textturnr Oka.xu\textturnr e]}) is incorrectly segmented into two separate words, viz., ``{\assamese{নৰক}}'' (\textipa{[nO\textturnr Ok]}'')and ``{\assamese{আশুৰে}}'' (\textipa{[axu\textturnr e]}),  disrupting the internal morphological structure of the word. The same pattern is seen for the word ``{\assamese{ দুৱাৰমুখত}}'' (\textipa{[duwa\textturnr muk\textsuperscript{H}Ot]}) in Sample 140. A boundary shift due to acoustic overlap of vowels occurred in Sample 71, replacing ``{\assamese{দৰা এতিয়াও}}'' (\textipa{[dO\textturnr a etijao]}) by ``{\assamese{ধৰাই তিয়াও}}'' (\textipa{[d\super{H}O\textturnr ai tijao]}). Another morphological error called \textit{lexical substitution} is seen in Sample 57, where meaningful phrase ``{\assamese{সুখ দুখ}}'' (\textipa{[xuk\textsuperscript{H} duk\textsuperscript{H}]}) (english meaning: \textit{happiness} and \textit{sorrow}) is replaced by the non existent word ``{\assamese{শুখটোক}}'' (\textipa{[xuk\textsuperscript{H}tUk]}). Additionally, affix-related errors, viz., suffix omission, addition, and substitution, are also frequently observed in the predicted transcripts. For instance, a suffix omission error is observed in Sample 12, in which the locative case marker `{\assamese{ত}}'{([t])} is lost in the predicted word ``{\assamese{কোলাত হেৰি}}'' (\textipa{[kUlat heri]}) for the reference word ``{\assamese{কলা ধেৰি}}'' (\textipa{[kOla d\super{H}e\.ri]}). In Sample 52, a suffix addition error is observed, where an unnecessary grapheme, `{\assamese{ম}}' (\textipa{/m/}), is appended to the root word ``{\assamese{কানাডালৈ}}'' (\textipa{[kana\textsubdot{d}alei]}). This type of error can be termed as grapheme over-generation problem.  Additionally, word distortion suffix substitution error is also observed in Sample 114, where the word ``{\assamese{আপোনালোকৰ}}'' (\textipa{[a.pUna.lU.kO\textturnr]}) is altered into the grammatically incorrect word, ``{\assamese{আপ্নলোকাৰ}}'' (\textipa{[apnO.lU.ka\textturnr]}). The vowel `{\assamese{ো}}' (\textipa{[U]}) in the root word ``{\assamese{আপোন}}'' (``\textipa{a.pUn}'') is deleted, and a conjunct character `{\assamese{প্ন}}' (\textipa{[pnO]}) is formed by combining the consonants, `{\assamese{প}}' (\textipa{[p]}) and `{\assamese{ন}}' (\textipa{[n]}). Additionally, an orthographical error is also observed in the form of the omission of the vowel `{\assamese{া}}' (\textipa{[a]}). 

The above findings indicate that accurate word boundary detection is crucial, and improving both tokenization and morpheme boundary identification is likely to reduce morphological errors, thereby contributing to a more robust and reliable ASR system.

\begin{table*}[t]
\caption{Morphological Error Analysis}
\label{MorphologicalErros}
\centering
\small
\renewcommand{\arraystretch}{1.2}

\begin{tabular}{p{1.2cm} p{4cm} p{4cm} p{6cm}}
\toprule
\textbf{Sample} & 
\textbf{REF Phrase} & 
\textbf{PRED Phrase} & 
\textbf{Morphological Errors Found} \\
\midrule

5 & {\assamese{সৰ্ব্ব প্ৰথম}} 
  & {\assamese{শ্বৰ্বপ্ৰথম}} 
  & Word Boundary Error (Merging) \\

5 & {\assamese{নৰকাসুৰে}} 
  & {\assamese{নৰক আশুৰে}} 
  & Word Boundary Error (Splitting) \\

12 & {\assamese{কোলাত হেৰি}} 
   & {\assamese{কলা ধেৰি}} 
   & Suffix Omission (Loss of case marker {\assamese{ত}}) \\

52 & {\assamese{কানাডালৈ}} 
   & {\assamese{কাণাডালৈম}} 
   & Suffix Addition (Overgeneration) \\

57 & {\assamese{সুখ দুখ}} 
   & {\assamese{শুখটোক}} 
   & Lexical Substitution \\

71 & {\assamese{দৰা এতিয়াও}} 
   & {\assamese{ধৰাই তিয়াও}} 
   & Word Boundary Error (Boundary Shifting) \\

114 & {\assamese{আপোনালোকৰ}} 
    & {\assamese{আপ্নলোকাৰ}} 
    & Suffix Substitution \\

140 & {\assamese{ছোৱালী কেজনী}} 
    & {\assamese{ছোৱালীকেইজনী}} 
    & Word Boundary Error (Merging) \\

140 & {\assamese{দুৱাৰমুখত}} 
    & {\assamese{দোৱাৰ মুখত}} 
    & Word Boundary Error (Splitting) \\

327 & {\assamese{কোৰআন শ্বৰীফ}} 
    & {\assamese{কুৰাণচৰিপ}} 
    & Word Boundary Error (Merging) + Lexical Substitution \\

\bottomrule
\end{tabular}
\end{table*}

\subsubsection{Correct Transcription Instances}
The Fine-tuned model encountered several challenges due to the complex morphology of the Assamese language and the need to adhere to standard orthographic patterns. Despite these difficulties, the model achieved remarkable success in several aspects. A Character Error Rate (CER) of $13.18\%$ demonstrates the ability of the model to effectively map acoustic signals to the corresponding script and 
generate orthographically correct transcripts. 
In contrast to the Zero-shot baseline, which failed to generate Assamese script, the fine-tuned model consistently produced it. The fine-tuned Whisper-Small successfully transcribed vowel-dense phonetic structure ``{\assamese{লৰাজনক সোধাত গম পালে তাৰ নাম পবিত্ৰ নাথ}}'' (\textipa{[lO\textturnr a.dZO.nOk xO.d\textsuperscript{H}at gOm pale tar nam pO.bi.t\textturnr O nat\textsuperscript{H}]}) as shown at Sample 41 in Table \ref{qualitative}. The model handles complex agglutinative morphology of the words like ``{\assamese{লৰাজনক}}'' (\textipa{[lO\textturnr a.dZO.nOk]}), consisting of the root word `{\assamese{লৰা}}' 
(\textipa{[lO.\textturnr a]} meaning `\textit{Boy}'), with a definitive suffix `{\assamese{জন}}' (\textipa{[dZOn]}) that transforms into ``\textit{The boy}'', which is further agglutinated with an accusative case marker `{\assamese{ক}}' (\textipa{[kO]}). Another successful instance is observed in Smaple 276, in which the whole reference script ``{\assamese{বিশেষকৈ শ্ৰীমান অভয় দুৱৰা আৰু শ্ৰীমান তৰুণ দুৱৰাৰ বাবেইহে}}'' (\textipa{[bi.xe.xO.kei x\textturnr i.mAn Ob\textsuperscript{H}Oj duwO\textturnr a a\textturnr u  x\textturnr i.man tO\textturnr un duwO\textturnr a\textturnr O          ba.be.i.He]}) is successfully transcribed. A similar instance of accurate and successful transcription is noted in sample 75 as well. Multi-syllable words like ``{\assamese{কাণ্ডকাৰখানা}}'' (\textipa{[kan.\textsubdot{d}O.ka\textturnr .k\textsuperscript{H}a.na]}), ``{\assamese{লভিতাহঁতে}}'' (\textipa{[lO.b\textsuperscript{H}i.ta.HOn.te]}), 
``{\assamese{মানুহটোৱে}}'' (\textipa{[ma.nuH.tU.we]}), and ``{\assamese{বিশেষকৈ}}'' (\textipa{[bi.xe.xO.kei]})   were handled elegantly by this model with limited or no error as demonstrated in Table \ref{qualitative}. Auxiliary and terminal verbs like ``{\assamese{গৈছিল}}'' (\textipa{[gOei.sil]}), ``{\assamese{কৰে}}'' (\textipa{[kO.re]}), and ``{\assamese{খালেগৈ}}'' (\textipa{[k\textsuperscript{H}a.le.gOei]}) found in Samples 52, 57 and 140, respectively, and are correctly understood by the model. High-frequency Assamese words such as ``{\assamese{ঘৰৰ}}'' (\textipa{[g\textsuperscript{H}O.\textturnr O\textturnr O]}), ``{\assamese{পৰিয়ালৰ}}'' (\textipa{[pO.\textturnr i.ja.lO\textturnr O]}), ``{\assamese{ইছলাম}}'' (\textipa{[islam]}), ``{\assamese{ধৰ্মৰ}}'' (\textipa{[d\textsuperscript{H}O\textturnr .mO\textturnr O]}), ``{\assamese{কৰিছিল}}'' (\textipa{[kO.ri.sil]}), ``{\assamese{হয়}}'' (\textipa{[HOj]}) were also transcribed without any error. Despite a relatively high Word Error Rate (WER), the Character Error Rate (CER) remains sufficiently low to preserve the semantic meaning of most sentences. Another notable milestone achieved by the model is its ability to correctly handle several conjunct consonants (juktaxar) in Assamese script. For instance, conjuncts viz., `{\assamese{ন্দ}}' (`{\assamese{ন}}' +  `{\assamese{দ}}'), and `{\assamese{স্ত}}' (`{\assamese{স}}' +  `{\assamese{ত}}') in Sample 114 (Table \ref{qualitative}) are correctly recognized in the word ``{\assamese{বন্দৱস্ত}}'' (\textipa{[bOn.dO.wOs.tO]}). 

Analysis of the fine-tuned model's transcripts clarifies quantitative reductions in MER and WIL, demonstrating the model's ability to achieve semantic alignment while preserving information. Qualitative analysis of the transcripts reveals that, despite a relatively high WER, the generated sentences maintain high readability and semantic coherence. This claim is further supported by METEOR and BLEU scores achieved by the model. Additionally, qualitative analysis of the fine-tuned Whisper-Small reveals that the model is sufficiently optimized to capture phonetic cues from the acoustic signals. However, it exhibits limitations in understanding morphological boundaries and orthographic rules in agglutinative languages like Assamese. This observation suggests that, although the model demonstrates strong phonetic recognition capabilities, its implicit lexical and morphological modeling remains insufficient for precise transcription in low-resource linguistic settings. Therefore, further performance improvements may be achieved by integrating the acoustic model with explicit language models, rather than simply scaling the acoustic architecture.
    
\section{Conclusions and Future Works} \label{conclusion}
In this study, we have established a new performance benchmark for Assamese ASR through controlled fine-tuning of the Whisper-small architecture on the Common-Voice Corpus 24.0 Assamese subset. The results demonstrate the adaptability of open-source multilingual models to low-resource speech recognition tasks. To address hardware constraints, mixed-precision (FP16) training, reduced batch sizes, and gradient accumulation were implemented. The model achieved a stable convergence around $3000$ steps with AdamW optimiser. The final evaluation of the model yields a WER of $43.75$\% and a CER of $13.18$\%. A significant improvement of  $93.10$\% in CER over the zero-shot baseline is achieved. The model also achieved an $78.26$\% improvement in WER. The low CER value indicates that the model has efficiently adapted phonetic-to-grapheme conversion in Assamese scripts. The use of additional error-based metrics further validates the model's proficiency, with reductions in MER to $0.4300$ and WIL to $0.6481$ from $1.0$. The $30.51$\% gap between CER and WER results from the agglutinative nature and lack of lexicon in the Assamese language. Text quality measurements demonstrate significant improvements, with BLEU and METEOR scores increasing to $30.81$ and $0.5262$ from $0.0$ in the zero-shot evaluation, indicating better semantic and structural correspondence between the predicted and reference transcriptions. Improvement in precision, recall, and F1-score, rising from $0.0$ in the zero-shot setting to balanced values $0.5960$, $0.5698$, and $0.5826$, respectively, indicating enhanced reliability in token-level predictions. Furthermore, the hallucination rate fell significantly from $0.5552$ to $0.0183$. This indicates that the generated outputs are becoming more grounded. At the same time, the real-time factor (RTF) was enhanced from $0.2943$ to $0.1990$, demonstrating more efficient inference after fine-tuning. Qualitative analysis reveals that the model is phonetically robust, with limited errors in morphological boundaries and suffix mutations. The fine-tuned acoustic model is highly competent, but it lacks linguistic knowledge. This rendered the bottleneck, preventing further improvements in WER. The proposed optimized model effectively learned phoneme-to-grapheme mappings, including Assamese conjunct characters (Juktaxar), and establishes a scalable framework for integrating a language model to develop a more robust ASR system for the Assamese language.

Several study directions can be explored in future, based on the findings of this study. To address the discrepancy between WER and CER, future research will investigate integrating grammatical constraints into language models to reduce segmentation errors and suffix distortion. Furthermore, the use of larger Whisper variants with increased parameters will be explored to enhance overall system performance.
Finally, an attempt will also be made to curate a high-quality Assamese speech dataset to improve coverage and robustness of the Whisper-based Assamese ASR system.

\section{Limitations}\label{Limitations}
Despite the significant performance improvements enabled by controlled fine-tuning, several constraints remain that define the current performance limits of Assamese Automatic Speech Recognition (ASR). The model configuration chosen for this study was dictated by the availability of computational resources. The training and testing were carried out using GPU resources available through the Kaggle research platform. Given the memory and computational constraints of these shared-GPU settings, the Whisper-Small architecture was chosen over its larger counterparts, such as Whisper-Medium or Whisper-Large-v3. Although the use of resource-efficient tactics like mixed-precision (FP16) training and gradient accumulation supported stable optimization, the reduced parameter capacity restricts its capacity to capture long-range contextual dependencies and intricate linguistic patterns. Employing larger-scale models with enhanced representational capacity could potentially enhance transcription accuracy and robustness.

The model is evaluated on the publicly available \textit{Common Voice 24.0 Assamese dataset}, which primarily consists of read speech recorded in fairly controlled acoustic conditions. Consequently, the model trained may show diminished robustness when applied to spontaneous conversations, dialectal differences, or noisy real-world acoustic environments. Assamese language exhibits significant regional variation, and dialects from regions like Lower Assam may introduce phonetic differences that are not adequately represented in the training dataset. Regional variation in Assamese, especially the dialects among Lower Assam and Upper Assam people, is not adequately represented in the training data, potentially affecting the performance of Whisper in recognizing Assamese speech.

\bibliographystyle{cas-model2-names}

\bibliography{cas-refs}

\begin{thebibliography}{67}
\expandafter\ifx\csname natexlab\endcsname\relax\def\natexlab#1{#1}\fi
\providecommand{\url}[1]{\texttt{#1}}
\providecommand{\href}[2]{#2}
\providecommand{\path}[1]{#1}
\providecommand{\DOIprefix}{doi:}
\providecommand{\ArXivprefix}{arXiv:}
\providecommand{\URLprefix}{URL: }
\providecommand{\Pubmedprefix}{pmid:}
\providecommand{\doi}[1]{\href{http://dx.doi.org/#1}{\path{#1}}}
\providecommand{\Pubmed}[1]{\href{pmid:#1}{\path{#1}}}
\providecommand{\bibinfo}[2]{#2}
\ifx\xfnm\relax \def\xfnm[#1]{\unskip,\space#1}\fi
\bibitem[{Agarwalla and Sarma(2016)}]{agarwalla2016machine}
\bibinfo{author}{Agarwalla, S.}, \bibinfo{author}{Sarma, K.K.},
  \bibinfo{year}{2016}.
\newblock \bibinfo{title}{Machine learning based sample extraction for
  automatic speech recognition using dialectal assamese speech}.
\newblock \bibinfo{journal}{Neural Networks} \bibinfo{volume}{78},
  \bibinfo{pages}{97--111}.
\newblock \DOIprefix\doi{10.1016/j.neunet.2015.12.010}.
\bibitem[{Akera et~al.(2025)Akera, Nafula, Walukagga, Yiga, Quinn and
  Mwebaze}]{akera2025much}
\bibinfo{author}{Akera, B.}, \bibinfo{author}{Nafula, E.},
  \bibinfo{author}{Walukagga, P.}, \bibinfo{author}{Yiga, G.},
  \bibinfo{author}{Quinn, J.}, \bibinfo{author}{Mwebaze, E.},
  \bibinfo{year}{2025}.
\newblock \bibinfo{title}{How much speech data is necessary for asr in african
  languages? an evaluation of data scaling in kinyarwanda and kikuyu}.
\newblock \bibinfo{journal}{arXiv preprint arXiv:2510.07221}
  \DOIprefix\doi{10.48550/arXiv.2510.07221}.
\bibitem[{Ardila et~al.(2020)Ardila, Branson, Davis, Kohler, Meyer, Henretty,
  Morais, Saunders, Tyers and Weber}]{ardila2020common}
\bibinfo{author}{Ardila, R.}, \bibinfo{author}{Branson, M.},
  \bibinfo{author}{Davis, K.}, \bibinfo{author}{Kohler, M.},
  \bibinfo{author}{Meyer, J.}, \bibinfo{author}{Henretty, M.},
  \bibinfo{author}{Morais, R.}, \bibinfo{author}{Saunders, L.},
  \bibinfo{author}{Tyers, F.}, \bibinfo{author}{Weber, G.},
  \bibinfo{year}{2020}.
\newblock \bibinfo{title}{Common voice: A massively-multilingual speech
  corpus}, in: \bibinfo{booktitle}{Proceedings of the 12th Language Resources
  and Evaluation Conference (LREC 2020)}, \bibinfo{publisher}{European Language
  Resources Association (ELRA)}, \bibinfo{address}{Marseille, France}. pp.
  \bibinfo{pages}{4218--4222}.
\newblock \DOIprefix\doi{10.48550/arXiv.1912.06670}.
\bibitem[{Baevski et~al.(2020)Baevski, Zhou, Mohamed and
  Auli}]{baevski2020wav2vec}
\bibinfo{author}{Baevski, A.}, \bibinfo{author}{Zhou, Y.},
  \bibinfo{author}{Mohamed, A.}, \bibinfo{author}{Auli, M.},
  \bibinfo{year}{2020}.
\newblock \bibinfo{title}{Wav2vec 2.0: A framework for self-supervised learning
  of speech representations}.
\newblock \bibinfo{journal}{Advances in Neural Information Processing Systems}
  \bibinfo{volume}{33}, \bibinfo{pages}{12449--12460}.
\newblock \DOIprefix\doi{10.48550/arXiv.2006.11477}.
\bibitem[{Banerjee and Lavie(2005)}]{banerjee2005meteor}
\bibinfo{author}{Banerjee, S.}, \bibinfo{author}{Lavie, A.},
  \bibinfo{year}{2005}.
\newblock \bibinfo{title}{{METEOR}: An automatic metric for {MT} evaluation
  with improved correlation with human judgments}, in:
  \bibinfo{booktitle}{Proceedings of the {ACL} Workshop on Intrinsic and
  Extrinsic Evaluation Measures for Machine Translation and/or Summarization},
  \bibinfo{publisher}{Association for Computational Linguistics},
  \bibinfo{address}{Ann Arbor, Michigan}. pp. \bibinfo{pages}{65--72}.
\newblock \DOIprefix\doi{10.3115/1626355.1626389}.
\bibitem[{Bekarystankyzy et~al.(2025)Bekarystankyzy, Razaque and
  Mamyrbayev}]{bekarystankyzy2025integrated}
\bibinfo{author}{Bekarystankyzy, A.}, \bibinfo{author}{Razaque, A.},
  \bibinfo{author}{Mamyrbayev, O.}, \bibinfo{year}{2025}.
\newblock \bibinfo{title}{Integrated end-to-end multilingual method for
  low-resource agglutinative languages using cyrillic scripts}.
\newblock \bibinfo{journal}{Journal of Industrial Information Integration}
  \bibinfo{volume}{43}, \bibinfo{pages}{100750}.
\newblock \DOIprefix\doi{10.1016/j.jii.2024.100750}.
\bibitem[{Bhandari and Harit(2026)}]{bhandari2026post}
\bibinfo{author}{Bhandari, A.}, \bibinfo{author}{Harit, G.},
  \bibinfo{year}{2026}.
\newblock \bibinfo{title}{Post-{ASR} correction for low-resource {Rajasthani}
  language}.
\newblock \bibinfo{journal}{ACM Transactions on Asian and Low-Resource Language
  Information Processing} \DOIprefix\doi{10.1145/3793254}.
\bibitem[{Bharali and Kalita(2021)}]{bharali2015comparative}
\bibinfo{author}{Bharali, S.S.}, \bibinfo{author}{Kalita, S.K.},
  \bibinfo{year}{2021}.
\newblock \bibinfo{title}{A comparative study of different features for
  isolated spoken word recognition using {HMM} with reference to assamese
  language}.
\newblock \bibinfo{journal}{International Journal of Speech Technology}
  \bibinfo{volume}{18}, \bibinfo{pages}{673--684}.
\newblock \DOIprefix\doi{10.1007/s10772-015-9311-7}.
\bibitem[{Bhat and Strik(2025)}]{bhat2025two}
\bibinfo{author}{Bhat, C.}, \bibinfo{author}{Strik, H.}, \bibinfo{year}{2025}.
\newblock \bibinfo{title}{Two-stage data augmentation for improved asr
  performance for dysarthric speech}.
\newblock \bibinfo{journal}{Computers in Biology and Medicine}
  \bibinfo{volume}{189}, \bibinfo{pages}{109954}.
\newblock \DOIprefix\doi{10.1016/j.compbiomed.2025.109954}.
\bibitem[{Burling(2003)}]{burling2003northeastern}
\bibinfo{author}{Burling, R.}, \bibinfo{year}{2003}.
\newblock \bibinfo{title}{The tibeto-burman languages of northeastern india},
  in: \bibinfo{editor}{Thurgood, G.}, \bibinfo{editor}{LaPolla, R.J.} (Eds.),
  \bibinfo{booktitle}{The Sino-Tibetan Languages}. \bibinfo{edition}{1st} ed..
  \bibinfo{publisher}{Routledge}, \bibinfo{address}{London and New York}.
  Routledge Language Family Series. chapter~\bibinfo{chapter}{11}, pp.
  \bibinfo{pages}{169--191}.
\newblock \DOIprefix\doi{10.4324/9780203221051-20}.
\bibitem[{Chen et~al.(2023)Chen, Hsu and Chang}]{chen2023accelerating}
\bibinfo{author}{Chen, C.Y.}, \bibinfo{author}{Hsu, Y.H.},
  \bibinfo{author}{Chang, C.c.}, \bibinfo{year}{2023}.
\newblock \bibinfo{title}{Accelerating hakka speech recognition research and
  development using the whisper model}, in: \bibinfo{booktitle}{Proceedings of
  the 35th Conference on Computational Linguistics and Speech Processing
  (ROCLING 2023)}, pp. \bibinfo{pages}{367--370}.
\bibitem[{Chen et~al.(2024)Chen, Chu, Li and Kawahara}]{chen2024data}
\bibinfo{author}{Chen, J.}, \bibinfo{author}{Chu, C.}, \bibinfo{author}{Li,
  S.}, \bibinfo{author}{Kawahara, T.}, \bibinfo{year}{2024}.
\newblock \bibinfo{title}{Data selection using spoken language identification
  for low-resource and zero-resource speech recognition}, in:
  \bibinfo{booktitle}{2024 Asia Pacific Signal and Information Processing
  Association Annual Summit and Conference (APSIPA ASC)},
  \bibinfo{publisher}{IEEE}. pp. \bibinfo{pages}{1--6}.
\newblock \DOIprefix\doi{10.1109/APSIPAASC63619.2025.10848811}.
\bibitem[{Chen et~al.(2025)Chen, Tian, Peng, Yan, Yang and
  Watanabe}]{chen2025owls}
\bibinfo{author}{Chen, W.}, \bibinfo{author}{Tian, J.}, \bibinfo{author}{Peng,
  Y.}, \bibinfo{author}{Yan, B.}, \bibinfo{author}{Yang, C.H.H.},
  \bibinfo{author}{Watanabe, S.}, \bibinfo{year}{2025}.
\newblock \bibinfo{title}{{OWLS}: Scaling laws for multilingual speech
  recognition and translation models}.
\newblock \bibinfo{journal}{arXiv preprint arXiv:2502.10373}
  \DOIprefix\doi{10.48550/arXiv.2502.10373}.
\bibitem[{Culotta et~al.(2007)Culotta, Wick, Hall, Marzilli and
  McCallum}]{culotta2007canonicalization}
\bibinfo{author}{Culotta, A.}, \bibinfo{author}{Wick, M.L.},
  \bibinfo{author}{Hall, R.J.}, \bibinfo{author}{Marzilli, M.},
  \bibinfo{author}{McCallum, A.}, \bibinfo{year}{2007}.
\newblock \bibinfo{title}{Canonicalization of database records using adaptive
  similarity measures}, in: \bibinfo{booktitle}{Proceedings of the 13th ACM
  SIGKDD International Conference on Knowledge Discovery and Data Mining},
  \bibinfo{publisher}{Association for Computing Machinery}. pp.
  \bibinfo{pages}{201--209}.
\newblock \DOIprefix\doi{10.1145/1281192.1281217}.
\bibitem[{Das and Bhattacharjee(2024)}]{das2024assamese}
\bibinfo{author}{Das, H.C.}, \bibinfo{author}{Bhattacharjee, U.},
  \bibinfo{year}{2024}.
\newblock \bibinfo{title}{Assamese dialect identification using static and
  dynamic features from vowel}.
\newblock \bibinfo{journal}{Journal of Advances in Information Technology}
  \bibinfo{volume}{15}, \bibinfo{pages}{306--317}.
\newblock \DOIprefix\doi{10.12720/jait.15.2.306-321}.
\bibitem[{Davis et~al.(1952)Davis, Biddulph and Balashek}]{davis1952automatic}
\bibinfo{author}{Davis, K.H.}, \bibinfo{author}{Biddulph, R.},
  \bibinfo{author}{Balashek, S.}, \bibinfo{year}{1952}.
\newblock \bibinfo{title}{Automatic recognition of spoken digits}.
\newblock \bibinfo{journal}{The Journal of the Acoustical Society of America}
  \bibinfo{volume}{24}, \bibinfo{pages}{637--642}.
\newblock \DOIprefix\doi{10.1121/1.1906946}.
\bibitem[{Davis and D{\"u}rst(2001)}]{davis2001unicode}
\bibinfo{author}{Davis, M.}, \bibinfo{author}{D{\"u}rst, M.},
  \bibinfo{year}{2001}.
\newblock \bibinfo{title}{Unicode Normalization Forms}.
\newblock \bibinfo{type}{Unicode Standard Annex \#15}. Unicode Consortium.
\newblock \URLprefix \url{https://www.unicode.org/reports/tr15/}.
\bibitem[{De~Vries et~al.(2014)De~Vries, Davel, Badenhorst, Basson, De~Wet,
  Barnard and De~Waal}]{de2014smartphone}
\bibinfo{author}{De~Vries, N.J.}, \bibinfo{author}{Davel, M.H.},
  \bibinfo{author}{Badenhorst, J.}, \bibinfo{author}{Basson, W.D.},
  \bibinfo{author}{De~Wet, F.}, \bibinfo{author}{Barnard, E.},
  \bibinfo{author}{De~Waal, A.}, \bibinfo{year}{2014}.
\newblock \bibinfo{title}{A smartphone-based asr data collection tool for
  under‑resourced languages}.
\newblock \bibinfo{journal}{Speech Communication} \bibinfo{volume}{56},
  \bibinfo{pages}{119--131}.
\newblock \DOIprefix\doi{10.1016/j.specom.2013.07.001}.
\bibitem[{Deka et~al.(2018)Deka, Nirmala and
  Samudravijaya}]{deka2018development}
\bibinfo{author}{Deka, B.}, \bibinfo{author}{Nirmala, S.R.},
  \bibinfo{author}{Samudravijaya, K.}, \bibinfo{year}{2018}.
\newblock \bibinfo{title}{Development of assamese continuous speech recognition
  system}, in: \bibinfo{booktitle}{Proceedings of the 6th Workshop on Spoken
  Language Technologies for Under-Resourced Languages (SLTU 2018)},
  \bibinfo{publisher}{ISCA}. pp. \bibinfo{pages}{220--224}.
\newblock \DOIprefix\doi{10.21437/SLTU.2018-46}.
\bibitem[{Dutta et~al.(2022)Dutta, Choudhury and Barman}]{dutta2022assamese}
\bibinfo{author}{Dutta, D.}, \bibinfo{author}{Choudhury, R.D.},
  \bibinfo{author}{Barman, U.}, \bibinfo{year}{2022}.
\newblock \bibinfo{title}{Assamese speech-based vocabulary identification
  system using convolutional neural network}.
\newblock \bibinfo{journal}{International Journal of Computing and Digital
  Systems} \bibinfo{volume}{12}, \bibinfo{pages}{1191--1202}.
\newblock \DOIprefix\doi{10.12785/ijcds/120195}.
\bibitem[{El~Hannani et~al.(2021)El~Hannani, Errattahi, Salmam, Hain and
  Ouahmane}]{el2021evaluation}
\bibinfo{author}{El~Hannani, A.}, \bibinfo{author}{Errattahi, R.},
  \bibinfo{author}{Salmam, F.Z.}, \bibinfo{author}{Hain, T.},
  \bibinfo{author}{Ouahmane, H.}, \bibinfo{year}{2021}.
\newblock \bibinfo{title}{Evaluation of the effectiveness and efficiency of
  state-of-the-art features and models for automatic speech recognition error
  detection}.
\newblock \bibinfo{journal}{Journal of Big Data} \bibinfo{volume}{8},
  \bibinfo{pages}{5}.
\newblock \DOIprefix\doi{10.1186/s40537-020-00391-w}.
\bibitem[{Goyal et~al.(2017)Goyal, Doll{\'a}r, Girshick, Noordhuis, Wesolowski,
  Kyrola, Tulloch, Jia and He}]{goyal2017accurate}
\bibinfo{author}{Goyal, P.}, \bibinfo{author}{Doll{\'a}r, P.},
  \bibinfo{author}{Girshick, R.}, \bibinfo{author}{Noordhuis, P.},
  \bibinfo{author}{Wesolowski, L.}, \bibinfo{author}{Kyrola, A.},
  \bibinfo{author}{Tulloch, A.}, \bibinfo{author}{Jia, Y.},
  \bibinfo{author}{He, K.}, \bibinfo{year}{2017}.
\newblock \bibinfo{title}{Accurate, large minibatch sgd: Training imagenet in 1
  hour}.
\newblock \bibinfo{journal}{arXiv preprint arXiv:1706.02677}
  \DOIprefix\doi{10.48550/arXiv.1706.02677}.
\bibitem[{Greer et~al.(1982)Greer, Lowerre and Wilcox}]{greer1982acoustic}
\bibinfo{author}{Greer, K.}, \bibinfo{author}{Lowerre, B.},
  \bibinfo{author}{Wilcox, L.}, \bibinfo{year}{1982}.
\newblock \bibinfo{title}{Acoustic pattern matching and beam searching}, in:
  \bibinfo{booktitle}{ICASSP '82. IEEE International Conference on Acoustics,
  Speech, and Signal Processing}, \bibinfo{publisher}{IEEE}. pp.
  \bibinfo{pages}{1251--1254}.
\newblock \DOIprefix\doi{10.1109/ICASSP.1982.1171508}.
\bibitem[{Hanson and Pratt(1988)}]{hanson1988comparing}
\bibinfo{author}{Hanson, S.J.}, \bibinfo{author}{Pratt, L.Y.},
  \bibinfo{year}{1988}.
\newblock \bibinfo{title}{Comparing biases for minimal network construction
  with back-propagation}, in: \bibinfo{booktitle}{Advances in Neural
  Information Processing Systems 1}, \bibinfo{publisher}{Morgan Kaufmann}. pp.
  \bibinfo{pages}{177--185}.
\bibitem[{Javed et~al.(2022)Javed, Doddapaneni, Raman, Bhogale, Ramesh,
  Kunchukuttan, Kumar and Khapra}]{javed2022towards}
\bibinfo{author}{Javed, T.}, \bibinfo{author}{Doddapaneni, S.},
  \bibinfo{author}{Raman, A.}, \bibinfo{author}{Bhogale, K.S.},
  \bibinfo{author}{Ramesh, G.}, \bibinfo{author}{Kunchukuttan, A.},
  \bibinfo{author}{Kumar, P.}, \bibinfo{author}{Khapra, M.M.},
  \bibinfo{year}{2022}.
\newblock \bibinfo{title}{Towards building asr systems for the next billion
  users}, in: \bibinfo{booktitle}{Proceedings of the AAAI Conference on
  Artificial Intelligence}, pp. \bibinfo{pages}{10813--10821}.
\newblock \DOIprefix\doi{10.1609/aaai.v36i10.21327}.
\bibitem[{Ji et~al.(2023)Ji, Lee, Frieske, Yu, Su, Xu, Ishii, Bang, Madotto and
  Fung}]{ji2023survey}
\bibinfo{author}{Ji, Z.}, \bibinfo{author}{Lee, N.}, \bibinfo{author}{Frieske,
  R.}, \bibinfo{author}{Yu, T.}, \bibinfo{author}{Su, D.}, \bibinfo{author}{Xu,
  Y.}, \bibinfo{author}{Ishii, E.}, \bibinfo{author}{Bang, Y.J.},
  \bibinfo{author}{Madotto, A.}, \bibinfo{author}{Fung, P.},
  \bibinfo{year}{2023}.
\newblock \bibinfo{title}{Survey of hallucination in natural language
  generation}.
\newblock \bibinfo{journal}{ACM Computing Surveys} \bibinfo{volume}{55},
  \bibinfo{pages}{1--38}.
\newblock \DOIprefix\doi{10.1145/3571730}.
\bibitem[{Kalita et~al.(2022)Kalita, Borbora and Nath}]{kalita2022use}
\bibinfo{author}{Kalita, D.}, \bibinfo{author}{Borbora, K.A.},
  \bibinfo{author}{Nath, D.}, \bibinfo{year}{2022}.
\newblock \bibinfo{title}{Use of bidirectional long short term memory in spoken
  word detection with reference to the assamese language}.
\newblock \bibinfo{journal}{Indian Journal of Science and Technology}
  \bibinfo{volume}{15}, \bibinfo{pages}{1364--1371}.
\newblock \DOIprefix\doi{10.17485/IJST/v15i27.655}.
\bibitem[{Kapusta et~al.(2024)Kapusta, Dr{\v{z}}{\'\i}k, {\v{S}}teflovi{\v{c}}
  and Nagy}]{kapusta2024text}
\bibinfo{author}{Kapusta, J.}, \bibinfo{author}{Dr{\v{z}}{\'\i}k, D.},
  \bibinfo{author}{{\v{S}}teflovi{\v{c}}, K.}, \bibinfo{author}{Nagy, K.S.},
  \bibinfo{year}{2024}.
\newblock \bibinfo{title}{Text data augmentation techniques for word embeddings
  in fake news classification}.
\newblock \bibinfo{journal}{{IEEE} Access} \bibinfo{volume}{12},
  \bibinfo{pages}{31538--31550}.
\newblock \DOIprefix\doi{10.1109/ACCESS.2024.3369918}.
\bibitem[{Kaur et~al.(2021)Kaur, Singh and Kadyan}]{kaur2021automatic}
\bibinfo{author}{Kaur, J.}, \bibinfo{author}{Singh, A.},
  \bibinfo{author}{Kadyan, V.}, \bibinfo{year}{2021}.
\newblock \bibinfo{title}{Automatic speech recognition system for tonal
  languages: State-of-the-art survey}.
\newblock \bibinfo{journal}{Archives of Computational Methods in Engineering}
  \bibinfo{volume}{28}, \bibinfo{pages}{1039--1068}.
\newblock \DOIprefix\doi{10.1007/s11831-020-09414-4}.
\bibitem[{Koenecke et~al.(2024)Koenecke, Choi, Mei, Schellmann and
  Sloane}]{koenecke2024careless}
\bibinfo{author}{Koenecke, A.}, \bibinfo{author}{Choi, A.S.G.},
  \bibinfo{author}{Mei, K.X.}, \bibinfo{author}{Schellmann, H.},
  \bibinfo{author}{Sloane, M.}, \bibinfo{year}{2024}.
\newblock \bibinfo{title}{Careless whisper: Speech-to-text hallucination
  harms}, in: \bibinfo{booktitle}{Proceedings of the 2024 ACM Conference on
  Fairness, Accountability, and Transparency (FAccT '24)},
  \bibinfo{publisher}{ACM New York, NY, USA}. pp. \bibinfo{pages}{1672--1681}.
\newblock \DOIprefix\doi{10.1145/3630106.3658996}.
\bibitem[{Koudounas et~al.(2025)Koudounas, La~Quatra, Giollo, Siniscalchi and
  Baralis}]{koudounas2025hallucination}
\bibinfo{author}{Koudounas, A.}, \bibinfo{author}{La~Quatra, M.},
  \bibinfo{author}{Giollo, M.}, \bibinfo{author}{Siniscalchi, S.M.},
  \bibinfo{author}{Baralis, E.}, \bibinfo{year}{2025}.
\newblock \bibinfo{title}{Hallucination benchmark for speech foundation
  models}.
\newblock \bibinfo{journal}{arXiv preprint arXiv:2510.16567}
  \DOIprefix\doi{10.48550/arXiv.2510.16567}.
\bibitem[{Labied et~al.(2024)Labied, Belangour and
  Banane}]{labied2024assessing}
\bibinfo{author}{Labied, M.}, \bibinfo{author}{Belangour, A.},
  \bibinfo{author}{Banane, M.}, \bibinfo{year}{2024}.
\newblock \bibinfo{title}{Assessing speech-to-text translation quality: An
  overview of key metrics}, in: \bibinfo{booktitle}{Proceedings of the 2024
  International Conference on Decision Aid Sciences and Applications (DASA)},
  \bibinfo{publisher}{IEEE}, \bibinfo{address}{Casablanca, Morocco}. pp.
  \bibinfo{pages}{1--6}.
\newblock \DOIprefix\doi{10.1109/DASA63652.2024.10836447}.
\bibitem[{Lambamo et~al.(2023)Lambamo, Srinivasagan and
  Jifara}]{lambamo2022analyzing}
\bibinfo{author}{Lambamo, W.}, \bibinfo{author}{Srinivasagan, R.},
  \bibinfo{author}{Jifara, W.}, \bibinfo{year}{2023}.
\newblock \bibinfo{title}{Analyzing noise robustness of cochleogram and mel
  spectrogram features in deep learning based speaker recognition}.
\newblock \bibinfo{journal}{Applied Sciences} \bibinfo{volume}{13},
  \bibinfo{pages}{569}.
\newblock \DOIprefix\doi{10.3390/app13010569}.
\bibitem[{Leben(2018)}]{leben2018languages}
\bibinfo{author}{Leben, W.R.}, \bibinfo{year}{2018}.
\newblock \bibinfo{title}{Languages of the world}, in:
  \bibinfo{booktitle}{Oxford Research Encyclopedia of Linguistics}.
  \bibinfo{publisher}{Oxford University Press}.
\newblock \DOIprefix\doi{10.1093/acrefore/9780199384655.013.349}.
\bibitem[{Liang et~al.(2025)Liang, Khaw, Liew, Tan and Qin}]{liang2025towards}
\bibinfo{author}{Liang, X.}, \bibinfo{author}{Khaw, Y.M.J.},
  \bibinfo{author}{Liew, S.Y.}, \bibinfo{author}{Tan, T.P.},
  \bibinfo{author}{Qin, D.}, \bibinfo{year}{2025}.
\newblock \bibinfo{title}{Towards low-resource languages machine translation: A
  language-specific fine-tuning with lora for specialized large language
  models}.
\newblock \bibinfo{journal}{IEEE Access} , \bibinfo{pages}{46616 --
  46626}\DOIprefix\doi{10.1109/ACCESS.2025.3549795}.
\bibitem[{Liu et~al.(2024)Liu, Yang and Qu}]{liu2024exploration}
\bibinfo{author}{Liu, Y.}, \bibinfo{author}{Yang, X.}, \bibinfo{author}{Qu,
  D.}, \bibinfo{year}{2024}.
\newblock \bibinfo{title}{Exploration of whisper fine-tuning strategies for
  low-resource asr}.
\newblock \bibinfo{journal}{EURASIP Journal on Audio, Speech, and Music
  Processing} \bibinfo{volume}{2024}, \bibinfo{pages}{29}.
\newblock \DOIprefix\doi{10.1186/s13636-024-00349-3}.
\bibitem[{Loshchilov and Hutter(2019)}]{loshchilov2017decoupled}
\bibinfo{author}{Loshchilov, I.}, \bibinfo{author}{Hutter, F.},
  \bibinfo{year}{2019}.
\newblock \bibinfo{title}{Decoupled weight decay regularization}, in:
  \bibinfo{booktitle}{Proceedings of the 7th International Conference on
  Learning Representations (ICLR)}, \bibinfo{address}{New Orleans, LA, USA}.
\newblock \DOIprefix\doi{10.48550/arXiv.1711.05101}.
\bibitem[{Lowerre(1990)}]{lowerre1990harpy}
\bibinfo{author}{Lowerre, B.}, \bibinfo{year}{1990}.
\newblock \bibinfo{title}{The {HARPY} speech understanding system}, in:
  \bibinfo{editor}{Waibel, A.}, \bibinfo{editor}{Lee, K.F.} (Eds.),
  \bibinfo{booktitle}{Readings in Speech Recognition}.
  \bibinfo{publisher}{Morgan Kaufmann Publishers Inc.}, \bibinfo{address}{San
  Francisco, CA, USA}, pp. \bibinfo{pages}{576--586}.
\newblock \DOIprefix\doi{10.1016/b978-0-08-051584-7.50053-x}.
\bibitem[{Magalhães et~al.(2022)Magalhães, Vasconcelos, Fernandes, Cruz,
  Sampaio, de~Macêdo and da~Silva}]{magalhaes2022evaluation}
\bibinfo{author}{Magalhães, R.P.}, \bibinfo{author}{Vasconcelos, D.J.R.},
  \bibinfo{author}{Fernandes, G.S.}, \bibinfo{author}{Cruz, L.A.},
  \bibinfo{author}{Sampaio, M.X.}, \bibinfo{author}{de~Macêdo, J.A.F.},
  \bibinfo{author}{da~Silva, T.L.C.}, \bibinfo{year}{2022}.
\newblock \bibinfo{title}{Evaluation of automatic speech recognition
  approaches}.
\newblock \bibinfo{journal}{Journal of Information and Data Management}
  \bibinfo{volume}{13}, \bibinfo{pages}{303--318}.
\newblock \DOIprefix\doi{10.5753/jidm.2022.2514}.
\bibitem[{Mahanta(2012)}]{mahanta2012assamese}
\bibinfo{author}{Mahanta, S.}, \bibinfo{year}{2012}.
\newblock \bibinfo{title}{Assamese}.
\newblock \bibinfo{journal}{Journal of the International Phonetic Association}
  \bibinfo{volume}{42}, \bibinfo{pages}{217--224}.
\newblock \DOIprefix\doi{10.1017/S0025100312000096}.
\bibitem[{Malah(1979)}]{malah1979time}
\bibinfo{author}{Malah, D.}, \bibinfo{year}{1979}.
\newblock \bibinfo{title}{Time-domain algorithms for harmonic bandwidth
  reduction and time scaling of speech signals}.
\newblock \bibinfo{journal}{IEEE Transactions on Acoustics, Speech, and Signal
  Processing} \bibinfo{volume}{27}, \bibinfo{pages}{121--133}.
\newblock \DOIprefix\doi{10.1109/TASSP.1979.1163210}.
\bibitem[{Medhi and Talukdar(2015)}]{medhi2015isolated}
\bibinfo{author}{Medhi, B.}, \bibinfo{author}{Talukdar, P.H.},
  \bibinfo{year}{2015}.
\newblock \bibinfo{title}{Isolated assamese speech recognition using artificial
  neural network}, in: \bibinfo{booktitle}{2015 International Symposium on
  Advanced Computing and Communication (ISACC)}, \bibinfo{publisher}{IEEE}. pp.
  \bibinfo{pages}{141--148}.
\newblock \DOIprefix\doi{10.1109/ISACC.2015.7377331}.
\bibitem[{Moral(1997)}]{moral1997north}
\bibinfo{author}{Moral, D.}, \bibinfo{year}{1997}.
\newblock \bibinfo{title}{North-east india as a linguistic area}.
\newblock \bibinfo{journal}{Mon-Khmer Studies} \bibinfo{volume}{27},
  \bibinfo{pages}{43--54}.
\bibitem[{Morris et~al.(2004)Morris, Maier and Green}]{morris2004and}
\bibinfo{author}{Morris, A.C.}, \bibinfo{author}{Maier, V.},
  \bibinfo{author}{Green, P.D.}, \bibinfo{year}{2004}.
\newblock \bibinfo{title}{{From WER and RIL to MER and WIL: improved evaluation
  measures for connected speech recognition.}}, in:
  \bibinfo{booktitle}{Interspeech}, pp. \bibinfo{pages}{2765--2768}.
\newblock \DOIprefix\doi{doi: 10.21437/Interspeech.2004-668}.
\bibitem[{{Mozilla Foundation}(2026)}]{mozilla}
\bibinfo{author}{{Mozilla Foundation}}, \bibinfo{year}{2026}.
\newblock \bibinfo{title}{Common voice dataset}.
\newblock
  \bibinfo{howpublished}{\url{https://datacollective.mozillafoundation.org/}}.
\newblock \bibinfo{note}{Accessed: 7 Mar 2026}.
\bibitem[{Nguyen-Duc et~al.(2025)Nguyen-Duc, Nguyen, Nguyen-Ho-Nhat, Nguyen and
  Lee}]{nguyen2025comparative}
\bibinfo{author}{Nguyen-Duc, M.}, \bibinfo{author}{Nguyen, L.V.},
  \bibinfo{author}{Nguyen-Ho-Nhat, H.}, \bibinfo{author}{Nguyen, T.H.},
  \bibinfo{author}{Lee, O.J.}, \bibinfo{year}{2025}.
\newblock \bibinfo{title}{A comparative study of deep audio models for
  spectrogram-and waveform-based singfake detection}.
\newblock \bibinfo{journal}{IEEE Access} \bibinfo{volume}{13},
  \bibinfo{pages}{15312--15325}.
\newblock \DOIprefix\doi{10.1109/ACCESS.2025.3571728}.
\bibitem[{Papineni et~al.(2002)Papineni, Roukos, Ward and
  Zhu}]{papineni2002bleu}
\bibinfo{author}{Papineni, K.}, \bibinfo{author}{Roukos, S.},
  \bibinfo{author}{Ward, T.}, \bibinfo{author}{Zhu, W.J.},
  \bibinfo{year}{2002}.
\newblock \bibinfo{title}{Bleu: A method for automatic evaluation of machine
  translation}, in: \bibinfo{booktitle}{Proceedings of the 40th Annual Meeting
  of the Association for Computational Linguistics},
  \bibinfo{publisher}{Association for Computational Linguistics},
  \bibinfo{address}{Philadelphia, Pennsylvania, USA}. pp.
  \bibinfo{pages}{311--318}.
\newblock \DOIprefix\doi{10.3115/1073083.1073135}.
\bibitem[{Perezhohin et~al.(2024)Perezhohin, Santos, Costa, Peres and
  Castelli}]{perezhohin2024enhancing}
\bibinfo{author}{Perezhohin, Y.}, \bibinfo{author}{Santos, T.},
  \bibinfo{author}{Costa, V.}, \bibinfo{author}{Peres, F.},
  \bibinfo{author}{Castelli, M.}, \bibinfo{year}{2024}.
\newblock \bibinfo{title}{Enhancing automatic speech recognition: effects of
  semantic audio filtering on models performance}.
\newblock \bibinfo{journal}{IEEE Access} \bibinfo{volume}{12},
  \bibinfo{pages}{155136--155150}.
\newblock \DOIprefix\doi{10.1109/ACCESS.2024.3482970}.
\bibitem[{Polat et~al.(2024)Polat, Turan, Ko{\c{c}}ak and
  Ula{\c{s}}}]{polat2024implementation}
\bibinfo{author}{Polat, H.}, \bibinfo{author}{Turan, A.K.},
  \bibinfo{author}{Ko{\c{c}}ak, C.}, \bibinfo{author}{Ula{\c{s}}, H.B.},
  \bibinfo{year}{2024}.
\newblock \bibinfo{title}{Implementation of a whisper architecture-based
  turkish automatic speech recognition (asr) system and evaluation of the
  effect of fine-tuning with a low-rank adaptation (lora) adapter on its
  performance}.
\newblock \bibinfo{journal}{Electronics} \bibinfo{volume}{13},
  \bibinfo{pages}{4227}.
\newblock \DOIprefix\doi{10.3390/electronics13214227}.
\bibitem[{Prasad et~al.(2026)Prasad, Rao, Naidu et~al.}]{prasad2026asr}
\bibinfo{author}{Prasad, C.}, \bibinfo{author}{Rao, V.G.S.},
  \bibinfo{author}{Naidu, R.C.A.}, et~al., \bibinfo{year}{2026}.
\newblock \bibinfo{title}{An asr transformer-based model for kannada
  speech-to-text transcription}.
\newblock \bibinfo{journal}{Journal of Artificial Intelligence and Technology}
  \DOIprefix\doi{10.37965/jait.2026.0935}.
\bibitem[{Rabiner(1989)}]{rabiner1989tutorial}
\bibinfo{author}{Rabiner, L.R.}, \bibinfo{year}{1989}.
\newblock \bibinfo{title}{A tutorial on hidden markov models and selected
  applications in speech recognition}.
\newblock \bibinfo{journal}{Proceedings of the IEEE} \bibinfo{volume}{77},
  \bibinfo{pages}{257--286}.
\newblock \DOIprefix\doi{10.1109/5.18626}.
\bibitem[{Rabiner and Juang(2006)}]{rabiner2006speech}
\bibinfo{author}{Rabiner, L.R.}, \bibinfo{author}{Juang, B.H.},
  \bibinfo{year}{2006}.
\newblock \bibinfo{title}{Speech recognition, automatic: History}, in:
  \bibinfo{editor}{Brown, K.} (Ed.), \bibinfo{booktitle}{Encyclopedia of
  Language and Linguistics}. \bibinfo{edition}{2} ed..
  \bibinfo{publisher}{Elsevier}, pp. \bibinfo{pages}{806--819}.
\newblock \DOIprefix\doi{10.1016/B0-08-044854-2/00906-8}.
\bibitem[{Radford et~al.(2023)Radford, Kim, Xu, Brockman, McLeavey and
  Sutskever}]{radford2023robust}
\bibinfo{author}{Radford, A.}, \bibinfo{author}{Kim, J.W.},
  \bibinfo{author}{Xu, T.}, \bibinfo{author}{Brockman, G.},
  \bibinfo{author}{McLeavey, C.}, \bibinfo{author}{Sutskever, I.},
  \bibinfo{year}{2023}.
\newblock \bibinfo{title}{Robust speech recognition via large-scale weak
  supervision}, in: \bibinfo{booktitle}{International conference on machine
  learning}, \bibinfo{organization}{PMLR}. pp. \bibinfo{pages}{28492--28518}.
\newblock \DOIprefix\doi{10.48550/arXiv.2212.04356}.
\bibitem[{Saikia and Camilleri(2019)}]{saikia2019assamese}
\bibinfo{author}{Saikia, P.}, \bibinfo{author}{Camilleri, M.},
  \bibinfo{year}{2019}.
\newblock \bibinfo{title}{Assamese case alignment shifts in progress}, in:
  \bibinfo{booktitle}{Proceedings of the LFG 2019 Conference},
  \bibinfo{publisher}{CSLI Publications}, \bibinfo{address}{Canberra,
  Australia}. pp. \bibinfo{pages}{251--271}.
\bibitem[{Sarma et~al.(2017)Sarma, Saharia and Sharma}]{sarma2017development}
\bibinfo{author}{Sarma, H.}, \bibinfo{author}{Saharia, N.},
  \bibinfo{author}{Sharma, U.}, \bibinfo{year}{2017}.
\newblock \bibinfo{title}{Development and analysis of speech recognition
  systems for assamese language using {HTK}}.
\newblock \bibinfo{journal}{ACM Transactions on Asian and Low-Resource Language
  Information Processing (TALLIP)} \bibinfo{volume}{17},
  \bibinfo{pages}{7:1--7:14}.
\newblock \DOIprefix\doi{10.1145/3137055}.
\bibitem[{Sarma and Sarma(2012)}]{sarma2012segmentation}
\bibinfo{author}{Sarma, M.}, \bibinfo{author}{Sarma, K.K.},
  \bibinfo{year}{2012}.
\newblock \bibinfo{title}{Segmentation and classification of vowel phonemes of
  assamese speech using a hybrid neural framework}.
\newblock \bibinfo{journal}{Applied Computational Intelligence and Soft
  Computing} \bibinfo{volume}{2012}, \bibinfo{pages}{1--8}.
\newblock \DOIprefix\doi{10.1155/2012/871324}.
\bibitem[{Sarma and Sarma(2014)}]{sarma2014phoneme}
\bibinfo{author}{Sarma, M.}, \bibinfo{author}{Sarma, K.K.},
  \bibinfo{year}{2014}.
\newblock \bibinfo{title}{Phoneme-Based Speech Segmentation Using Hybrid Soft
  Computing Framework}. volume \bibinfo{volume}{550} of
  \textit{\bibinfo{series}{Studies in Computational Intelligence}}.
\newblock \bibinfo{publisher}{Springer International Publishing},
  \bibinfo{address}{Cham, Switzerland}.
\newblock \DOIprefix\doi{10.1007/978-81-322-1862-3}.
\bibitem[{Sarma and Sarma(2011)}]{sarma2011assamese}
\bibinfo{author}{Sarma, M.P.}, \bibinfo{author}{Sarma, K.K.},
  \bibinfo{year}{2011}.
\newblock \bibinfo{title}{Assamese numeral speech recognition using multiple
  features and cooperative lvq-architectures}.
\newblock \bibinfo{journal}{International Journal of Electrical and
  Electronics} \bibinfo{volume}{5}, \bibinfo{pages}{1}.
\bibitem[{Sarma(2025)}]{sarma2025intersections}
\bibinfo{author}{Sarma, V.M.}, \bibinfo{year}{2025}.
\newblock \bibinfo{title}{Intersections between heritage, multilingualism, and
  education: Language acquisition in india}.
\newblock \bibinfo{journal}{Frontiers in Human Neuroscience}
  \bibinfo{volume}{19}, \bibinfo{pages}{1538482}.
\newblock \DOIprefix\doi{10.3389/fnhum.2025.1538482}.
\bibitem[{Shahnawazuddin et~al.(2013)Shahnawazuddin, Thotappa, Sarma, Deka,
  Prasanna and Sinha}]{shahnawazuddin2013assamese}
\bibinfo{author}{Shahnawazuddin, S.}, \bibinfo{author}{Thotappa, D.},
  \bibinfo{author}{Sarma, B.D.}, \bibinfo{author}{Deka, A.},
  \bibinfo{author}{Prasanna, S.R.M.}, \bibinfo{author}{Sinha, R.},
  \bibinfo{year}{2013}.
\newblock \bibinfo{title}{Assamese spoken query system to access the price of
  agricultural commodities}, in: \bibinfo{booktitle}{2013 National Conference
  on Communications (NCC)}, \bibinfo{publisher}{IEEE}. pp.
  \bibinfo{pages}{1--5}.
\newblock \DOIprefix\doi{10.1109/NCC.2013.6488011}.
\bibitem[{Sharma et~al.(2025)Sharma, Pandya and Shukla}]{sharma2025fine}
\bibinfo{author}{Sharma, A.K.}, \bibinfo{author}{Pandya, M.},
  \bibinfo{author}{Shukla, A.}, \bibinfo{year}{2025}.
\newblock \bibinfo{title}{Fine-tuning whisper tiny for swahili asr: Challenges
  and recommendations for low-resource speech recognition}, in:
  \bibinfo{booktitle}{Proceedings of the Sixth Workshop on African Natural
  Language Processing (AfricaNLP 2025)}, \bibinfo{publisher}{Association for
  Computational Linguistics}, \bibinfo{address}{Vienna, Austria}. pp.
  \bibinfo{pages}{74--81}.
\newblock \DOIprefix\doi{10.18653/v1/2025.africanlp-1.11}.
\bibitem[{Singh et~al.(2023)Singh, Mehta, Nanavati, Bandekar, Basumatary,
  Badiger, Udupa, Kumar, Ghosh, Pai et~al.}]{singh2023model}
\bibinfo{author}{Singh, A.}, \bibinfo{author}{Mehta, A.S.},
  \bibinfo{author}{Nanavati, J.}, \bibinfo{author}{Bandekar, J.},
  \bibinfo{author}{Basumatary, K.}, \bibinfo{author}{Badiger, S.},
  \bibinfo{author}{Udupa, S.}, \bibinfo{author}{Kumar, S.},
  \bibinfo{author}{Ghosh, P.K.}, \bibinfo{author}{Pai, P.}, et~al.,
  \bibinfo{year}{2023}.
\newblock \bibinfo{title}{Model adaptation for asr in low-resource indian
  languages}.
\newblock \bibinfo{journal}{arXiv preprint arXiv:2307.07948}
  \DOIprefix\doi{10.48550/arXiv.2307.07948}.
\bibitem[{Singh et~al.(2024)Singh, Zhong, Wang, Mendes, Hasegawa-Johnson,
  Abdulla and Shahamiri}]{singh2024comprehensive}
\bibinfo{author}{Singh, S.}, \bibinfo{author}{Zhong, Z.},
  \bibinfo{author}{Wang, Q.}, \bibinfo{author}{Mendes, C.},
  \bibinfo{author}{Hasegawa-Johnson, M.}, \bibinfo{author}{Abdulla, W.},
  \bibinfo{author}{Shahamiri, S.R.}, \bibinfo{year}{2024}.
\newblock \bibinfo{title}{A comprehensive performance evaluation of whisper
  models in dysarthric speech recognition}, in: \bibinfo{booktitle}{Neural
  Information Processing. ICONIP 2024}, \bibinfo{publisher}{Springer},
  \bibinfo{address}{Singapore}. pp. \bibinfo{pages}{75--90}.
\newblock \DOIprefix\doi{10.1007/978-981-96-6960-8_6}.
\bibitem[{Sourav et~al.(2025)Sourav, Javaid and Cheng}]{sourav2025review}
\bibinfo{author}{Sourav, M.S.G.}, \bibinfo{author}{Javaid, A.},
  \bibinfo{author}{Cheng, L.}, \bibinfo{year}{2025}.
\newblock \bibinfo{title}{A review of ai in human‑machine cooperation:
  Machine perspective}.
\newblock \bibinfo{journal}{ACM Transactions on Autonomous and Adaptive
  Systems} \bibinfo{volume}{20}, \bibinfo{pages}{1--24}.
\newblock \DOIprefix\doi{10.1145/3774318}.
\bibitem[{Timmel et~al.(2024)Timmel, Paonessa, Vogel, Perruchoud and
  Kakooee}]{timmel2025fine}
\bibinfo{author}{Timmel, V.}, \bibinfo{author}{Paonessa, C.},
  \bibinfo{author}{Vogel, M.}, \bibinfo{author}{Perruchoud, D.},
  \bibinfo{author}{Kakooee, R.}, \bibinfo{year}{2024}.
\newblock \bibinfo{title}{Fine-tuning whisper on low-resource languages for
  real-world applications}.
\newblock \bibinfo{journal}{arXiv preprint arXiv:2412.15726}
  \DOIprefix\doi{10.48550/arXiv.2412.15726}.
\bibitem[{Yu and Deng(2015)}]{yu2015automatic}
\bibinfo{author}{Yu, D.}, \bibinfo{author}{Deng, L.}, \bibinfo{year}{2015}.
\newblock \bibinfo{title}{Automatic Speech Recognition: A Deep Learning
  Approach}. volume~\bibinfo{volume}{1} of \textit{\bibinfo{series}{Signals and
  Communication Technology}}.
\newblock \bibinfo{publisher}{Springer}, \bibinfo{address}{London}.
\newblock \DOIprefix\doi{10.1007/978-1-4471-5779-3}.
\bibitem[{de~Zuazo et~al.(2025)de~Zuazo, Navas, Saratxaga and
  Rioja}]{de2025whisper}
\bibinfo{author}{de~Zuazo, X.}, \bibinfo{author}{Navas, E.},
  \bibinfo{author}{Saratxaga, I.}, \bibinfo{author}{Rioja, I.H.},
  \bibinfo{year}{2025}.
\newblock \bibinfo{title}{Whisper-lm: Improving asr models with language models
  for low-resource languages}.
\newblock \bibinfo{journal}{arXiv preprint arXiv:2503.23542}
  \DOIprefix\doi{10.48550/arXiv.2503.23542}.

\end{thebibliography}



\end{document}